\documentclass[11pt]{article}

\usepackage[]{acl}
\usepackage{comment}
\usepackage{booktabs}
\usepackage{pifont}
\newcommand{\cmark}{\ding{51}}

\usepackage{booktabs}
\usepackage{graphicx}

\usepackage{tikz}
\usetikzlibrary{arrows.meta,positioning,fit,backgrounds,calc}
\usepackage{graphicx}

\usepackage[most]{tcolorbox}

\usepackage{times}
\usepackage{latexsym}
\usepackage{multirow}

\usepackage[T1]{fontenc}
\usepackage[utf8]{inputenc}
\usepackage{CJKutf8} % local support for Chinese examples only

\usepackage{microtype}

\usepackage{inconsolata}

\usepackage{graphicx}

\title{Mind the Gap: Exposing LLM Translation Blind Spots \\ Using the AlphaMWE Multilingual Parallel Corpus}

  \author{ Lifeng Han$^{1,2}$, Jiahui Liang$^1$,  
  Anna Latusek$^3$,  
  Karim El Haff$^4$,\\
  \textbf{Amal Haddad Haddad}$^5$,
  \textbf{Josua Höfgen}$^6$,
  \textbf{Kilian Evang}$^7$, 
  \textbf{Min Ma}$^8$, 
  \textbf{Maryia Zhyrko}$^1$ \\
  $^1$\texttt{Leiden University} 
  $^2$\texttt{Leiden University Medical Centre} \\
   $^3$\texttt{Institute of Computer Science PAS, Warsaw} \\
   $^4$\texttt{Faculty of Computer and Information Science, University of Ljubljana}\\
   $^5$\texttt{University of Granada}
   $^6$\texttt{
Technical University of Munich
}\\
   $^7$\texttt{Heinrich Heine University Düsseldorf}    $^8$\texttt{Google DeepMind} \\
j.h.l.jiahui@hum.leidenuniv.nl | l.han@lumc.nl
 \\
 \\
}

\begin{document}
\begin{CJK*}{UTF8}{gbsn}
\maketitle
\begin{abstract}
% This document is a supplement to the general instructions for *ACL authors. It contains instructions for using the \LaTeX{} style files for ACL conferences.
% The document itself conforms to its own specifications, and is therefore an example of what your manuscript should look like.
% These instructions should be used both for papers submitted for review and for final versions of accepted papers.
LLMs' performance on machine translation (MT) tasks is often dependent on the data availability in the specific domains and language pairs that they are trained upon.
To examine if multiword expressions (MWEs) still present a bottleneck for LLMs regarding language understanding and translation, 
we report the performance of systems from the WMT 2026 Test Suites shared task, using portions of the publicly available multilingual parallel corpus AlphaMWE as test suites.
We received 31 MT systems' outputs covering English to Chinese (zh), Polish (pl), German (de), and Arabic (ar) including Modern Standard Arabic (MSA) and two dialectal ones (Egyptian and Tunisian Arabic).
We carried out automatic evaluations using BLEU, ChrF, and BERT-score to select the top 3 systems per language pair, followed up with human evaluations on the selected systems.
Our findings show that figurative and MWE-related phenomena remain challenging for contemporary MT systems, automatic metrics sometimes disagree in system ranking, and human evaluation uncovers language-specific errors that remain hidden by aggregate scores. Inter-annotator analysis further reveals challenges in consistently identifying and calibrating linguistically subtle translation errors, highlighting the need for explicit evaluation guidelines and careful annotator calibration.\footnote{Accepted by WMT2026, to be presented at 28-29 October, 2026
Budapest, Hungary \url{https://www2.statmt.org/wmt26/}}
% Our findings show that: figurative/MWE phenomena remain challenging; automatic metrics sometimes disagree; human evaluation uncovers language-specific errors hidden by aggregate scores.
\end{abstract}

\section{Introduction}

Large Language Models (LLMs) have substantially advanced machine
translation (MT), achieving remarkable fluency and strong automatic
evaluation scores across many language pairs. Nevertheless, recent
studies suggest that high overall translation quality does not imply
that all linguistic phenomena are equally well handled. In particular,
multiword expressions (MWEs) such as idioms, metaphorical language, and other
non-compositional constructions remain challenging because their
meaning cannot always be inferred compositionally from individual
words. Correct translation therefore requires not only lexical
knowledge but also contextual reasoning, semantic interpretation, and
language-specific linguistic competence.

Recent work has begun to investigate these remaining blind spots.
\citet{cheng-amiri-2025-linguistic} argue that LLMs exhibit systematic linguistic
blind spots despite their strong overall performance. \citet{laureano-de-leon-etal-2025-evaluating} demonstrate that multilingual and code-switched MWEs
remain difficult for current LLMs, while \citet{mi-etal-2025-rolling} show that
idiomatic expressions are frequently mistranslated when successful
interpretation depends on contextual understanding rather than surface
word associations. More recently, \citet{liu-etal-2026-evaluating,liang2026metahope} reported that
verbal MWEs and metaphors continue to pose substantial challenges for machine
translation systems. Together, these studies suggest that automatic MT
quality improvements have not eliminated linguistically difficult
translation phenomena.

However, most previous work evaluates LLMs in controlled experimental
settings or focuses on individual linguistic phenomena. Less is known
about how a diverse collection of contemporary MT systems behaves on a
shared multilingual benchmark specifically designed to expose these
challenging constructions. Furthermore, automatic evaluation metrics
such as BLEU or BERTScore often provide only an overall quality
estimate and may overlook linguistically important translation errors,
particularly those involving idiomaticity, lexical ambiguity, or
figurative meaning.

To investigate these issues, we participated in the WMT 2026 Test
Suites Shared Task by submitting AlphaMWE, a multilingual parallel
corpus containing manually annotated multiword expressions, as a
diagnostic test suite. AlphaMWE covers English--Chinese,
English--German, English--Polish, and English--Arabic (Modern Standard
Arabic, Egyptian Arabic, and Tunisian Arabic), and contains texts from
multiple genres including technical documentation, news, and
literature. The shared task attracted 31 MT systems, providing a broad
comparison of contemporary translation models.

Our evaluation follows a three-stage pipeline. We first construct the
WMT-compatible test suites from AlphaMWE, then evaluate all submitted
systems using SacreBLEU, chrF++, and BERTScore, and finally perform
HOPE-based human evaluation on the top-ranked systems for each
language pair. This combination enables us to compare automatic and
human assessment while analysing translation behaviour across multiple
languages and linguistic phenomena.

% \textcolor{red}
{It is worth noting that, at this stage of our work, we decided to conduct the analysis and comparisons at the level of complete sentences. We therefore assessed the meaning of each sentence as a whole, rather than focusing solely on the translations of the MWEs. Thus, our primary focus was on the overall meaning conveyed by sentences containing MWEs, as well as on identifying which types of errors resulting from the use of MWEs occur in different types of texts.}

Our analysis shows that despite strong automatic evaluation scores,
idioms, verbal MWEs, metaphorical expressions, lexical ambiguity, and
language-specific collocations remain important sources of translation
errors. These shortcomings are especially evident in literary and
expression-rich texts, whereas technical texts are translated much more reliably. We further observe that automatic evaluation metrics do
not always agree on system ranking, and that human evaluation reveals
language-specific translation issues which remain largely invisible to
aggregate automatic scores.
\newline
\newline 
Our main contributions are:

\begin{itemize}
\item We introduce AlphaMWE as a multilingual diagnostic test suite in
the WMT 2026 Test Suites Shared Task, covering six target-language
variants and outputs from 31 participating MT systems.

\item We combine three complementary automatic evaluation metrics with
HOPE-based human evaluation to investigate both system ranking and
linguistically motivated translation quality.

\item We provide a cross-language analysis showing that MWEs,
idiomatic expressions, figurative language, lexical ambiguity, and
language-specific grammatical phenomena remain persistent blind spots
for contemporary MT systems despite their overall high translation
quality.
\end{itemize}

\section{WMT2026 Test Suites: AlphaMWE}

% \begin{figure*}[t]
%   \includegraphics[width=.99\textwidth]{Figs/3stages.png}
%   \caption{Stages of the Shared Task}
%   \label{fig:3stages}
% \end{figure*}

% Add to the preamble:
% \usepackage{tikz}
% \usetikzlibrary{arrows.meta,positioning,calc,backgrounds}
% \usepackage{graphicx}

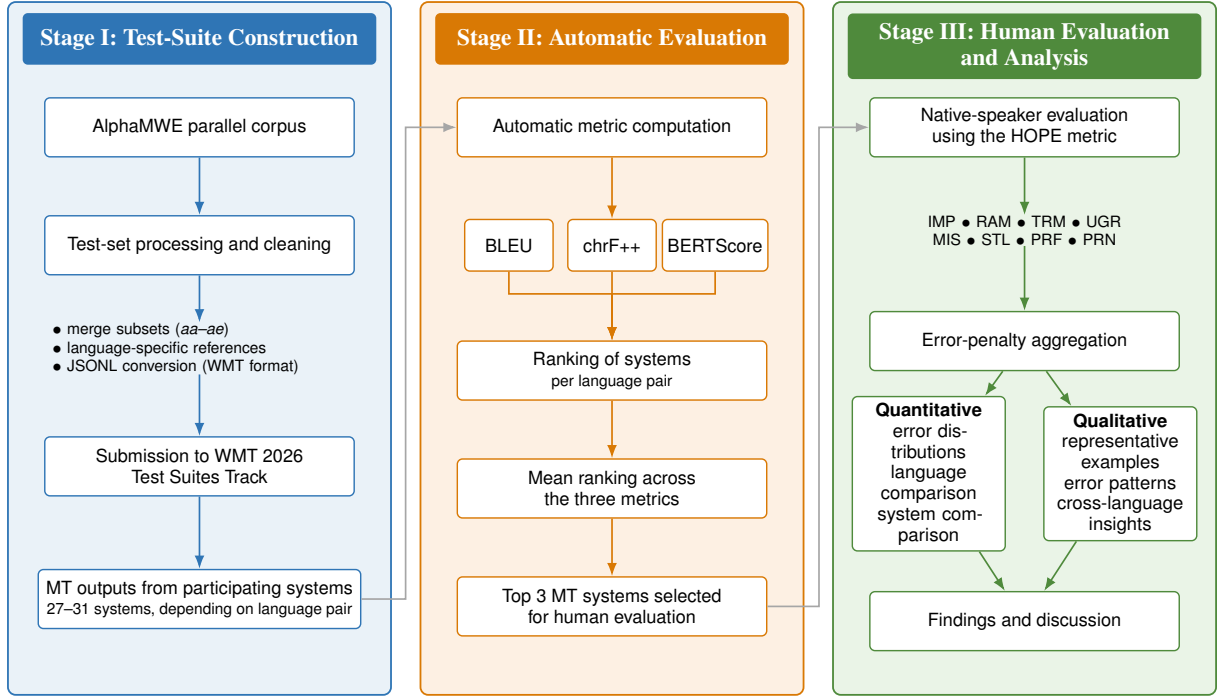
\begin{figure*}[t]
\centering
\resizebox{\textwidth}{!}{%
\begin{tikzpicture}[
    font=\sffamily\scriptsize,
    >=Latex,
    main/.style={
        draw,
        rounded corners=2pt,
        align=center,
        fill=white,
        minimum width=42mm,
        minimum height=8mm,
        inner sep=3pt,
        line width=0.55pt
    },
    metric/.style={
        draw,
        rounded corners=2pt,
        align=center,
        fill=white,
        minimum width=12mm,
        minimum height=7mm,
        inner sep=2pt,
        line width=0.55pt
    },
    analysis/.style={
        draw,
        rounded corners=2pt,
        align=center,
        fill=white,
        text width=19mm,
        minimum height=8mm,
        inner sep=2.5pt,
        line width=0.55pt
    },
    arrow/.style={-{Latex[length=1.8mm]}, line width=0.65pt},
    title/.style={
        text=white,
        font=\bfseries\footnotesize,
        rounded corners=2pt,
        minimum width=48mm,
        minimum height=7mm,
        align=center
    },
    note/.style={
        align=left,
        text width=40mm,
        inner sep=0pt,
        font=\sffamily\tiny
    }
]

\definecolor{blue}{HTML}{2F75B5}
\definecolor{bluefill}{HTML}{EAF2F8}
\definecolor{orange}{HTML}{D97904}
\definecolor{orangefill}{HTML}{FDF0E3}
\definecolor{green}{HTML}{4F8A3C}
\definecolor{greenfill}{HTML}{EAF4E7}

% Fixed stage backgrounds: 52 mm wide, with 4 mm gaps.
\begin{scope}[on background layer]
  \fill[bluefill,rounded corners=3pt]   (-26mm,-86mm) rectangle (26mm,8mm);
  \draw[blue,rounded corners=3pt,line width=0.7pt] (-26mm,-86mm) rectangle (26mm,8mm);

  \fill[orangefill,rounded corners=3pt] (30mm,-86mm) rectangle (82mm,8mm);
  \draw[orange,rounded corners=3pt,line width=0.7pt] (30mm,-86mm) rectangle (82mm,8mm);

  \fill[greenfill,rounded corners=3pt]  (86mm,-86mm) rectangle (138mm,8mm);
  \draw[green,rounded corners=3pt,line width=0.7pt] (86mm,-86mm) rectangle (138mm,8mm);
\end{scope}

% =========================================================
% Stage I
% =========================================================
\node[title,fill=blue] at (0,3mm) {Stage I: Test-Suite Construction};

\node[main,draw=blue] (corpus) at (0,-9mm)
{AlphaMWE parallel corpus};

\node[main,draw=blue] (processing) at (0,-25mm)
{Test-set processing and cleaning};

\node[note] (processingnote) at (0,-39mm)
{$\bullet$ merge subsets (\textit{aa--ae})\\
 $\bullet$ language-specific references\\
 $\bullet$ JSONL conversion (WMT format)};

\node[main,draw=blue] (submission) at (0,-55mm)
{Submission to WMT 2026\\Test Suites Track};

\node[main,draw=blue] (outputs) at (0,-73mm)
{MT outputs from participating systems\\
{\tiny 27--31 systems, depending on language pair}};

\draw[arrow,blue] (corpus) -- (processing);
\draw[arrow,blue] (processing) -- (processingnote);
\draw[arrow,blue] (processingnote) -- (submission);
\draw[arrow,blue] (submission) -- (outputs);

% =========================================================
% Stage II
% =========================================================
\node[title,fill=orange] at (56mm,3mm)
{Stage II: Automatic Evaluation};

\node[main,draw=orange] (metrics) at (56mm,-9mm)
{Automatic metric computation};

\node[metric,draw=orange] (bleu) at (42mm,-25mm) {BLEU};
\node[metric,draw=orange] (chrf) at (56mm,-25mm) {chrF++};
\node[metric,draw=orange] (bert) at (70mm,-25mm) {BERTScore};

\node[main,draw=orange] (rank) at (56mm,-42mm)
{Ranking of systems\\{\tiny per language pair}};

\node[main,draw=orange] (meanrank) at (56mm,-58mm)
{Mean ranking across\\the three metrics};

\node[main,draw=orange] (topthree) at (56mm,-74mm)
{Top 3 MT systems selected\\for human evaluation};

\draw[arrow,orange] (metrics) -- (chrf);
\draw[arrow,orange] (bleu.south) -- ++(0,-3mm) -| (rank.north);
\draw[arrow,orange] (chrf.south) -- (rank.north);
\draw[arrow,orange] (bert.south) -- ++(0,-3mm) -| (rank.north);
\draw[arrow,orange] (rank) -- (meanrank);
\draw[arrow,orange] (meanrank) -- (topthree);

% =========================================================
% Stage III
% =========================================================
\node[title,fill=green] at (112mm,2mm)
{Stage III: Human Evaluation\\ and Analysis
};

\node[main,draw=green] (human) at (112mm,-9mm)
{Native-speaker evaluation\\using the HOPE metric};

\node[note,align=center,text width=42mm] (errors) at (112mm,-23mm)
{IMP $\bullet$ RAM $\bullet$ TRM $\bullet$ UGR\\
 MIS $\bullet$ STL $\bullet$ PRF $\bullet$ PRN};

\node[main,draw=green] (aggregation) at (112mm,-38mm)
{Error-penalty aggregation};

\node[analysis,draw=green] (quant) at (99mm,-56mm)
{\textbf{Quantitative}\\
error distributions\\
language comparison\\
system comparison};

\node[analysis,draw=green] (qual) at (125mm,-56mm)
{\textbf{Qualitative}\\
representative examples\\
error patterns\\
cross-language insights};

\node[main,draw=green] (findings) at (112mm,-76mm)
{Findings and discussion};

\draw[arrow,green] (human) -- (errors);
\draw[arrow,green] (errors) -- (aggregation);
\draw[arrow,green] (aggregation) -- (quant);
\draw[arrow,green] (aggregation) -- (qual);
\draw[arrow,green] (quant) -- (findings);
\draw[arrow,green] (qual) -- (findings);

% Cross-stage arrows
\draw[arrow,gray!70] (outputs.east) -- ++(6mm,0) |- (metrics.west);
\draw[arrow,gray!70] (topthree.east) -- ++(7mm,0) |- (human.west);

\end{tikzpicture}%
}
\caption{Overall methodology for AlphaMWE test-suite construction, automatic evaluation of WMT 2026 system outputs, top 3 system selection per language pair, and HOPE-based human evaluation followed by quantitative and qualitative analyses.}
\label{fig:overall-methodology}
\end{figure*}

We present the overall methodology of this shared task in Figure \ref{fig:overall-methodology} including three stages:
Stage I: Test Suite Construction; Stage II: Automatic Evaluation; Stage III: Human Evaluation \& Analysis.

For Stage I, AlphaMWE \cite{han2026towards} is a multilingual parallel corpus featuring multiword expression (MWE) annotations, machine translation post-editing, and human-in-the-loop quality controls. It builds on data from sources such as the PARSEME shared task \cite{parseme2017shared} to evaluate how translation engines handle complex phrasal and verbal expressions.
% AlphaMWE is a Multilingual parallel corpus with MWE annotations \cite{han2026towards}. 
It was originally designed by \newcite{han-etal-2020-alphamwe} for English--German, English--Polish, and English--Chinese language pairs, which was later extended to the Arabic edition by \newcite{hadj-mohamed-etal-2023-alphamwe}.
AlphaMWE has a mix of text genres, including IT document instructions, news/politics, and literature texts.

The original AlphaMWE corpus contained 5 files of equal size, with around 150 segments per file, for a total of 747 segments. However, their annotation was not complete for English--Arabic and English--Polish, resulting in 150 segments for English--MSA (modern standard Arabic), English--AEB (Tunisian Arabic), English--ARZ (Egyptian Arabic), and 597 segments for English--Polish\footnote{Following ISO 639-3 language identifiers.}.

For Stage II, we deployed SacreBLEU \cite{post-2018-sacrebleu}, chrF++ \cite{popovic-2017-chrf}, and BERTScore \cite{zhang2020bertscore} for precision-oriented n-gram matching, robust character-level matching, and contextual semantic similarity.

For system selection,
% ranking:
% Generating tables:
we apply average rank instead of average score.
Average rank is preferable because BLEU, chrF++, and BERTScore have different numerical scales.
We rank the system per language pair using each metric, then average their ranking on all tested metrics.

For Stage III, we adopt the human-centric post-editing based metric HOPE \cite{gladkoff-han-2022-hope} with its original eight error types and severity levels (0, 1, 2, 4, 8, 16) for (minor, medium, major, severe, and critical).
The eight error types are: Impact (IMP), Required Adaptation Missing (RAM), Terminology (TRM), Ungrammatical (UGR), Mistranslation (MIS), Style (STL), Proofreading (PRF), and Proper Name (PRN).

\section{Systems Received}
We list the statistics of MT system outputs we received from WMT in Table \ref{tab:wmt_system_coverage}. The varying numbers of participating systems reflect the fact that not all WMT 2026 Test Suites participants submitted translations for every language pair.

\begin{table*}[t]
\centering
\small
\caption{Coverage of MT systems that returned translations for the AlphaMWE test suites submitted to the WMT~2026 Test Suites Shared Task. Numbers in parentheses indicate the number of evaluation segments in each language pair. English--Polish contains 597 segments because the \textit{ad} subset was unavailable, while the Arabic test suites each contain the 150-segment \textit{aa} subset only.}
\label{tab:wmt_system_coverage}
\begin{tabular}{lcccccc}
\toprule
\textbf{System}
& \textbf{en$\rightarrow$aeb}
& \textbf{en$\rightarrow$ar}
& \textbf{en$\rightarrow$arz}
& \textbf{en$\rightarrow$de}
& \textbf{en$\rightarrow$pl}
& \textbf{en$\rightarrow$zh} \\
& \textbf{(150)}
& \textbf{(150)}
& \textbf{(150)}
& \textbf{(747)}
& \textbf{(597)}
& \textbf{(747)} \\
\midrule
% \toprule
% \textbf{System}
% & \textbf{en$\rightarrow$aeb\\(150)}
% & \textbf{en$\rightarrow$ar\\(150)}
% & \textbf{en$\rightarrow$arz\\(150)}
% & \textbf{en$\rightarrow$de\\(747)}
% & \textbf{en$\rightarrow$pl\\(597)}
% & \textbf{en$\rightarrow$zh\\(747)} \\
% \midrule
CUNI-AR            & \cmark & \cmark & \cmark &        &        &        \\
CUNI-EdUKate       &        &        &        & \cmark &        &        \\
CUNI-MH-v3         &        &        &        & \cmark &        &        \\
Cohere CAT+        & \cmark & \cmark & \cmark & \cmark & \cmark & \cmark \\
Command A+         & \cmark & \cmark & \cmark & \cmark & \cmark & \cmark \\
DeepSeek V4 Pro    & \cmark & \cmark & \cmark & \cmark & \cmark & \cmark \\
Dubformer          & \cmark & \cmark & \cmark & \cmark & \cmark & \cmark \\
GLM 4--9B          & \cmark & \cmark & \cmark & \cmark & \cmark & \cmark \\
GPT 5.5            & \cmark & \cmark & \cmark & \cmark & \cmark & \cmark \\
GPT OSS 120B       & \cmark & \cmark & \cmark & \cmark & \cmark & \cmark \\
GPT OSS 20B        & \cmark & \cmark & \cmark & \cmark & \cmark & \cmark \\
Gemini 3.1 Pro     & \cmark & \cmark & \cmark & \cmark & \cmark & \cmark \\
Gemma 4--31B       & \cmark & \cmark & \cmark & \cmark & \cmark & \cmark \\
Gemma 4--4B        & \cmark & \cmark & \cmark & \cmark & \cmark & \cmark \\
Google Translate   & \cmark & \cmark & \cmark & \cmark & \cmark & \cmark \\
HW-TSC             &        & \cmark &        &        &        & \cmark \\
Lumen              &        & \cmark & \cmark & \cmark & \cmark & \cmark \\
Ministral 3--14B   & \cmark & \cmark & \cmark & \cmark & \cmark & \cmark \\
Mistral Medium 3.5 & \cmark & \cmark & \cmark & \cmark & \cmark & \cmark \\
Qingqiu-MT-9B      & \cmark & \cmark & \cmark & \cmark & \cmark & \cmark \\
Qwen 3.5--9B       & \cmark & \cmark & \cmark & \cmark & \cmark & \cmark \\
Qwen 3.6--27B      & \cmark & \cmark & \cmark & \cmark & \cmark & \cmark \\
SCIR-TG-MT         &        &        &        &        &        & \cmark \\
SRPOL              &        &        & \cmark & \cmark &        & \cmark \\
SalamandraTA       & \cmark & \cmark & \cmark & \cmark & \cmark & \cmark \\
TRIVE              & \cmark & \cmark & \cmark & \cmark & \cmark & \cmark \\
Tiny Aya Global    &        & \cmark & \cmark & \cmark & \cmark & \cmark \\
Tower 9B           & \cmark & \cmark & \cmark & \cmark & \cmark & \cmark \\
UvA-MT             &        &        & \cmark & \cmark &        & \cmark \\
VoxNexus\_V1       &        & \cmark & \cmark & \cmark &        &        \\
Wayfinder          &        & \cmark & \cmark & \cmark & \cmark & \cmark \\
\midrule
\textbf{\# Systems}
& \textbf{24}
& \textbf{28}
& \textbf{29}
& \textbf{31}
& \textbf{27}
& \textbf{27} \\
\bottomrule
\end{tabular}
\end{table*}

\begin{table*}[th!]
\centering
% \scriptsize
\small
\setlength{\tabcolsep}{3.5pt}
\renewcommand{\arraystretch}{1.08}
\caption{Top 3 MT systems selected for human evaluation on each AlphaMWE language pair.
Systems were selected according to the average of their SacreBLEU, chrF++, and BERTScore F1 rankings.
The rank standard deviation (Rank SD) measures the agreement among the three automatic metrics; lower values indicate greater agreement but were \emph{not} used as a selection criterion. Competition ranking was adopted for the final selection rank.}
\label{tab:human_eval_top3}
\begin{tabular}{llrrrrrr}
\toprule
\textbf{Language Pair} &
\textbf{System} &
\textbf{BLEU} &
\textbf{chrF++} &
\textbf{BERT} &
\textbf{Avg. Rank} &
\textbf{Rank SD} &
\textbf{Selection Rank} \\
\midrule

en$\rightarrow$aeb &
TRIVE &
1 &
1 &
1 &
1.000 &
0.000 &
1 \\

&
CUNI-AR &
2 &
2 &
2 &
2.000 &
0.000 &
2 \\

&
DeepSeek V4 Pro &
4 &
4 &
4 &
4.000 &
0.000 &
3 \\\hline 

\addlinespace

en$\rightarrow$ar &
GoogleTranslate &
1 &
1 &
1 &
1.000 &
0.000 &
1 \\

&
Gemma 4--4B &
2 &
2 &
2 &
2.000 &
0.000 &
2 \\

&
TRIVE &
3 &
3 &
3 &
3.000 &
0.000 &
3 \\\hline 

\addlinespace

en$\rightarrow$arz &
Wayfinder &
2 &
1 &
3 &
2.000 &
0.816 &
1 \\

&
Gemma 4--31B &
5 &
5 &
1 &
3.667 &
1.886 &
2 \\

&
GPT 5.5 &
1 &
2 &
9 &
4.000 &
3.559 &
3 \\
\hline 
\addlinespace

en$\rightarrow$de &
VoxNexus\_V1 &
1 &
1 &
1 &
1.000 &
0.000 &
1 \\

&
DeepSeek V4 Pro &
2 &
2 &
2 &
2.000 &
0.000 &
2 \\

&
TRIVE &
3 &
3 &
3 &
3.000 &
0.000 &
3 \\
\hline 
\addlinespace

en$\rightarrow$pl &
DeepSeek V4 Pro &
1 &
1 &
1 &
1.000 &
0.000 &
1 \\

&
TRIVE &
3 &
2 &
2 &
2.333 &
0.471 &
2 \\

&
Dubformer &
2 &
3 &
3 &
2.667 &
0.471 &
3 \\
\hline 
\addlinespace

en$\rightarrow$zh &
SCIR-TG-MT &
2 &
2 &
1 &
1.667 &
0.471 &
1 \\

&
GoogleTranslate &
1 &
1 &
3 &
1.667 &
0.943 &
1 \\

&
Tower 9B &
3 &
4 &
2 &
3.000 &
0.816 &
3 \\

\bottomrule
\end{tabular}
\end{table*}
\section{Automatic Evaluation}

The top 3 systems per translation language pair are displayed in Table \ref{tab:human_eval_top3}, where we presented the metrics of BLEU, chrF++, BERT, average rank (Avg. Rank), rank standard deviation (Rank SD), and the selection rank.
%
% \textcolor{red}
% {This fragment looks as it was a note; are we going to keep it or maybe we need to elaborate the first part?: "}
Regarding the rank standard deviation (Rank-SD),
a low standard deviation indicates that all three automatic metrics agree on the system's quality, while a high standard deviation highlights disagreement between metrics.\footnote{Score 0 → all three metrics ranked the system identically (very high agreement).
Small SD (0.4–1.0) → often used for strong agreement.
Large SD (>2) → the metrics disagree substantially, suggesting that the system's quality depends on the evaluation metric.}
% This can be useful when discussing why a system was selected for human evaluation.

The top 3 systems for each language pair were selected based on the mean of the three automatic metric rankings. 
The rank standard deviation (SD) is reported to indicate the level of agreement among the automatic metrics but was not used as a selection criterion.
Nevertheless, we can see from the table that while half of the selected systems across language pairs have Rank-SD value 0 reflecting their stable performances across three metrics, there are another half of the systems having different level of Rank-SD, with the extremist from GPT 5.5 and Gemma 4-31B on the en→arz language pair having SD values of 3.559 and 1.886, very unbalanced ranking across metric.
Appendix Section \ref{sec:auto-scores-sys-langs} shows the three metric-specific ranking tables for transparency and reproducibility.

For Chinese, the competition-ranking tie is preserved:
SCIR-TG-MT: aggregate rank 1;
GoogleTranslate: aggregate rank 1;
Tower 9B: aggregate rank 3.
Here, the SD immediately explains the tie for en→zh: both the first two systems have the same mean rank (1.667).
SCIR-TG-MT has a lower SD (0.471), meaning the three metrics agree more closely.
Google Translate has a higher SD (0.943), indicating more disagreement because BERTScore ranked it lower than BLEU and chrF++.
Even though the official competition ranking remains 1, 1, 3, the SD provides valuable context for interpreting that tie.

% \begin{figure*}[t]
%   \includegraphics[width=\textwidth]{Figs/all-sys-langs-original-scores.png}
%   \caption{All MT systems on language pairs, original scores}
%   \label{fig:all-sys-langs-scores}
% \end{figure*}

% \begin{figure*}[t]
%   \includegraphics[width=\textwidth]{Figs/all-sys-langs-ranked.png}
%   \caption{All MT systems on language pairs, ranked}
%   \label{fig:all-sys-langs-ranked}
% \end{figure*}

\section{Human Evaluation}
\begin{figure*}[t]
  \includegraphics[width=.99\textwidth]{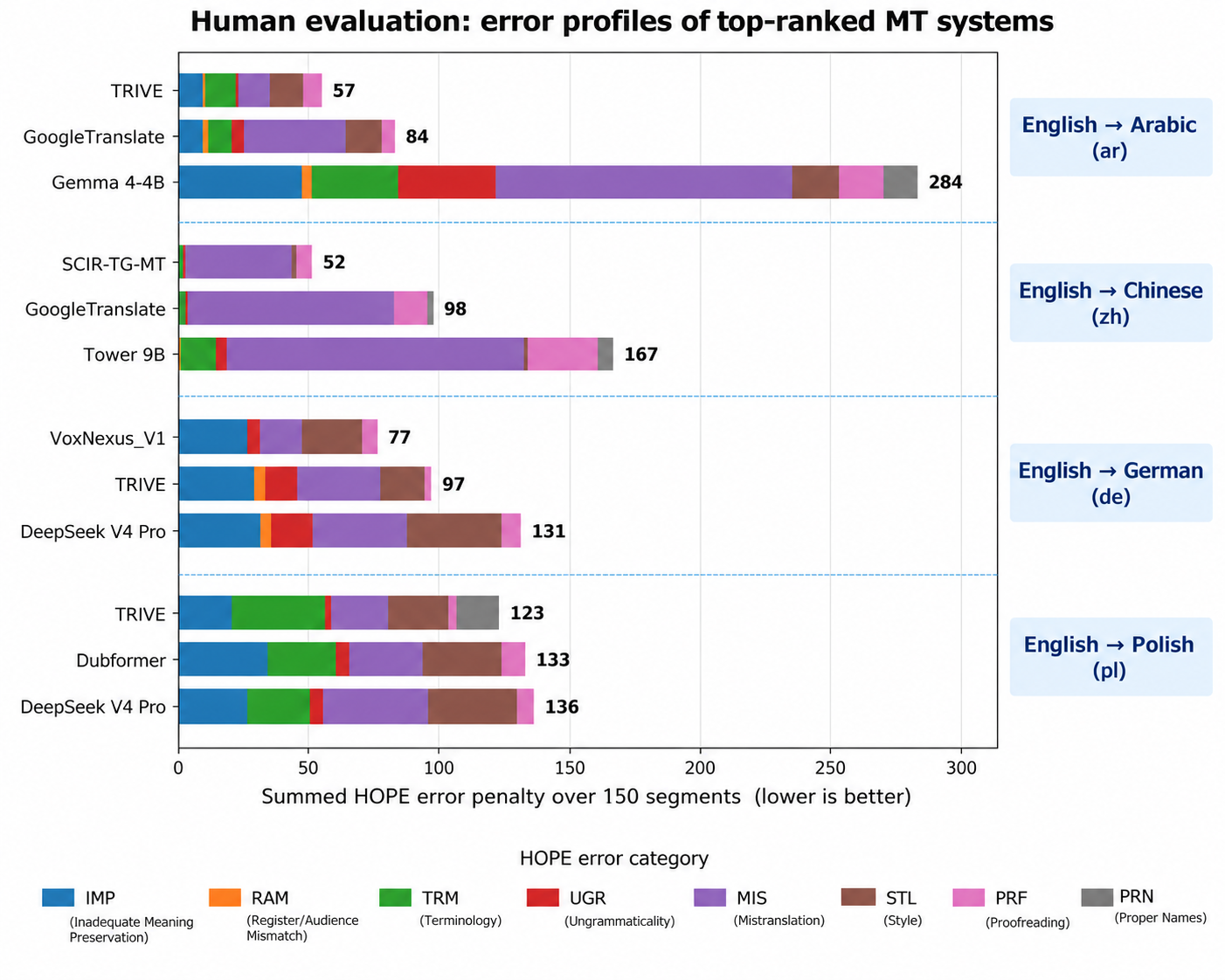}
  \caption{Visualisation of Human Evaluations on Four Language Pairs.}
  \label{fig:overall-eval-human-visual}
\end{figure*}

In Table \ref{tab:hope-error-penalty-totals}, we list the summed HOPE error penalties for the top 3 systems per language pair over 150 segments.
Because every language pair contains exactly 150 segments, these raw totals are directly comparable in terms of corpus size. However, possible differences in annotator severity across target languages should still be acknowledged when making cross-language interpretations.
Figure \ref{fig:overall-eval-human-visual} visualises the HOPE-based human evaluation results. Notably, the human-evaluation ranking does not fully reproduce the automatic-evaluation ranking of the selected top-three systems: the ordering changes for English–Arabic, English–German, and English–Polish, while human evaluation resolves the automatic-ranking tie between SCIR-TG-MT and Google Translate for English–Chinese.

% The visualisation of human evaluation is shown in Figure \ref{fig:overall-eval-human-visual}, where we can see that the order of quality ranking of selected Top-3 systems per language pair has changed, in comparison to automatic evaluation outputs.

\begin{table*}[t]
\centering
\small
\setlength{\tabcolsep}{6pt}
\renewcommand{\arraystretch}{1.08}
\caption{Summed HOPE error penalties for the three systems selected for human evaluation on each language pair. Each value is the total penalty assigned to an error category over 150 AlphaMWE segments. Lower scores indicate fewer or less severe translation errors. \textsc{Segs} is the sum of the eight error-category penalties.}
\label{tab:hope-error-penalty-totals}
\begin{tabular}{llrrrrrrrrr}
\toprule
\textbf{Language Pair} &
\textbf{MT System} &
\textbf{IMP} &
\textbf{RAM} &
\textbf{TRM} &
\textbf{UGR} &
\textbf{MIS} &
\textbf{STL} &
\textbf{PRF} &
\textbf{PRN} &
\textbf{SEGS} \\
\midrule

\multirow{3}{*}{en$\rightarrow$zh}
& SCIR-TG-MT       & 0  & 0 & 2  & 1  & 41  & 2  & 6  & 0  & 52  \\
& GoogleTranslate  & 0  & 0 & 3  & 1  & 79  & 0  & 13 & 2  & 98  \\
& Tower 9B         & 0  & 2 & 13 & 4  & 113 & 2  & 27 & 6  & 167 \\\hline
\addlinespace[3pt]

\multirow{3}{*}{en$\rightarrow$pl}
& DeepSeek V4 Pro  & 27 & 0 & 24 & 5  & 40  & 34 & 6  & 0  & 136 \\
& TRIVE            & 21 & 0 & 36 & 2  & 22  & 23 & 3  & 16 & 123 \\
& Dubformer        & 35 & 0 & 26 & 5  & 28  & 30 & 9  & 0  & 133 \\\hline
\addlinespace[3pt]

\multirow{3}{*}{en$\rightarrow$de}
& VoxNexus\_V1     & 27 & 0 & 0  & 5  & 16  & 23 & 6  & 0  & 77  \\
& DeepSeek V4 Pro  & 32 & 4 & 0  & 16 & 36  & 36 & 7  & 0  & 131 \\
& TRIVE            & 30 & 4 & 0  & 12 & 32  & 17 & 2  & 0  & 97  \\\hline
\addlinespace[3pt]

\multirow{3}{*}{en$\rightarrow$ar}
& GoogleTranslate  & 10 & 2 & 9  & 5  & 39  & 14 & 5  & 0  & 84  \\
& Gemma 4--4B      & 48 & 4 & 33 & 37 & 114 & 18 & 17 & 13 & 284 \\
& TRIVE            & 10 & 1 & 12 & 1  & 12  & 13 & 8  & 0  & 57  \\

\bottomrule
\end{tabular}
\end{table*}

\subsection{English--Arabic}
For Arabic, error annotations across the evaluated sentences revealed substantial differences in translation quality among the three machine translation systems. 
% Using the 2nd half of the 150 segments:
% TRIVE achieved the strongest overall performance, with errors identified in only 5 of 75 sentences and a cumulative severity score of 6, no instance exceeding moderate severity. 
% Google Translate exhibited an intermediate profile, with errors present in 19 sentences and a cumulative severity score of 34. 
% Gemma 4-4B showed the weakest performance, with errors identified in 31 sentences and a cumulative severity score of 72, more than double that of Google Translate and twelve times that of TRIVE.
% on 150 segs bellow:
TRIVE achieved the strongest overall performance, with errors identified in fewer sentences and a cumulative severity score of 57 on all segments. %6, no instance exceeding moderate severity. 
Google Translate exhibited an intermediate profile, with errors present in more sentences and a cumulative severity score of 84 on all 150 segments, mostly due to MIS error type over TRIVE. 
Gemma 4-4B showed the weakest performance, with errors identified in even more sentences and a cumulative severity score of 284, more than the sum of Google Translate and TRIVE, also introducing many errors in Impact category, followed by Terminology, Style, and Proofreading.

\begin{figure}[t]
  \centering
  \fcolorbox{gray!30}{gray!5}{%
    \parbox{.45\textwidth}{%
      \centering
      \includegraphics[width=.4\textwidth]{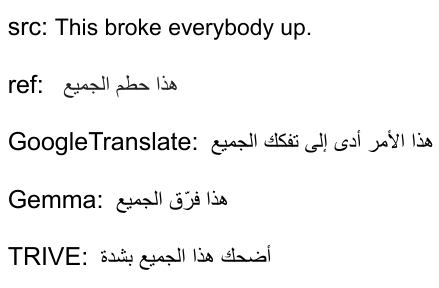}
      
      \vspace{2mm}
      
      \caption{Example on Arabic Translation Issues with vMWE (aa-105).}
      \label{fig:arabic-example-TRIVE-right}
    }%
  }

\end{figure}

Across all three systems, Mistranslation constituted the single largest error category, indicating that semantic accuracy, rather than syntactic well-formedness, was the primary locus of quality degradation. 
Ungrammatical output was comparatively rare; the few genuine grammar violations identified involved gender agreement failures, such as a masculine verb form incorrectly applied to a feminine subject. 
More frequently, apparent grammatical irregularities on closer inspection reflected stylistic or collocational choices rather than true violations of Arabic grammar, underscoring the importance of distinguishing genuine ungrammaticality from register mismatch or reduced naturalness in translation evaluation.
Several recurring error patterns emerged at the lexical and idiomatic level. Idiomatic expressions posed a particular challenge: an English idiom conveying collective laughter was rendered literally by two systems, producing a semantically unrelated and ultimately incoherent Arabic sentence, while the third system correctly recovered the idiomatic sense. For instance, in the following example in Figure \ref{fig:arabic-example-TRIVE-right}, TRIVE is the only system that correctly conveyed the meaning of laughter in the expression “this broke everybody up”, the other outputs convey the literal meaning of the breaking (fracturing) of ‘everyone'.

%\begin{figure}[t]
  %\includegraphics[width=.4\textwidth]{Figs/sent-id-aa105-ar_cropped2.pdf}
  %\caption{Example on Arabic Translation Issues with vMWE (sentence-id: aa-105).}
  %\label{fig:arabic-example-TRIVE-right}
%\end{figure}

% It is sentence 105

% src: This broke everybody up.

% ref:        هذا حطم الجميع

% GoogleTranslate:        هذا الأمر أدى إلى تفكك الجميع

% Gemma:      هذا فرّق الجميع

% TRIVE :     أضحك هذا الجميع بشدة

% TRIVE is the only system that correctly conveyed the meaning of laughter in the expression “this broke everybody up”, the other outputs convey the literal meaning of the breaking (fracturing) of ‘everyone'.

Similarly, polysemous source terms were occasionally mapped to the wrong sense entirely, as when a medical usage of a common English noun was translated according to its more frequent, unrelated meaning, yielding an implausible reading within its surrounding context. 
Errors of this kind, in which an incorrect lexical sense generates a locally incoherent target sentence, were judged more severe than errors of nuance, since they compromise not only precision but basic interpretability for the target reader.
Terminological consistency also varied across systems. In several instances, an established or institutionally fixed Arabic term used correctly elsewhere in the same output was replaced by a looser or less standard alternative, producing internal inconsistency within a single document. 
Related issues included inconsistent transliteration of proper names, occasional calques that preserved English syntactic structure at the expense of natural Arabic phrasing, and isolated failures to adapt source-language conventions, such as untranslated units of measurement or an uncorrected error present in the source text itself, appropriately treated as a required-adaptation failure when the target system failed to resolve it.

Taken together, these findings suggest that translation adequacy in this sample was governed less by surface grammaticality, which was generally well preserved across all systems, than by the accurate resolution of idiomatic expressions, lexical ambiguity, and terminological consistency. 
TRIVE demonstrated the most robust handling of these higher-order challenges, while Gemma 4-4B was disproportionately affected by mistranslation and inconsistent register, with Google Translate occupying an intermediate position. 

\subsection{English--Chinese}

\begin{figure}[t]
  \centering
  \fcolorbox{gray!30}{gray!5}{%
    \parbox{.45\textwidth}{%
      \centering
      \includegraphics[width=.4\textwidth]{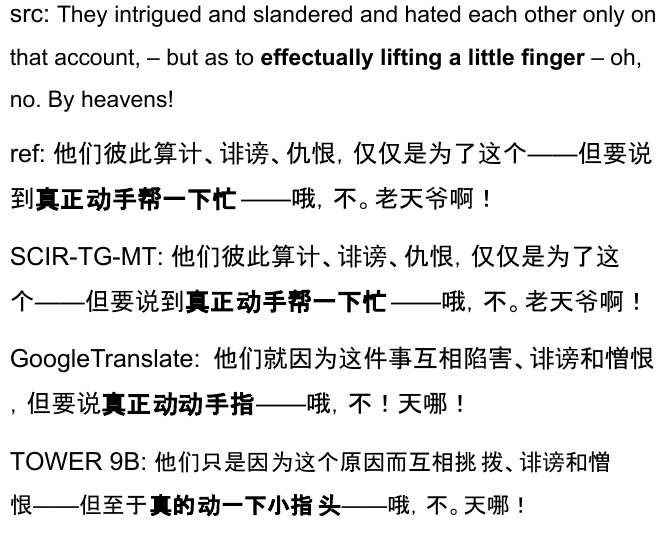}
      
      \vspace{2mm}
      
      \caption{Example on English-Chinese Translation Issues of  Metaphors (aa-127).}
      \label{fig:sent-id-127-en-zh}
    }%
  }
\end{figure}

%\begin{figure}[t]
  %\includegraphics[width=.48\textwidth]{Figs/en-zh-example-aa127_cropped.pdf}
  %\caption{Example on English-Chinese Translation Issues of  Metaphors (sentence-id: aa-127).}
  %\label{fig:sent-id-127-en-zh}
%\end{figure}

\begin{figure}[t]
  \centering
  \fcolorbox{gray!30}{gray!5}{%
    \parbox{.45\textwidth}{%
      \centering
      \includegraphics[width=.4\textwidth]{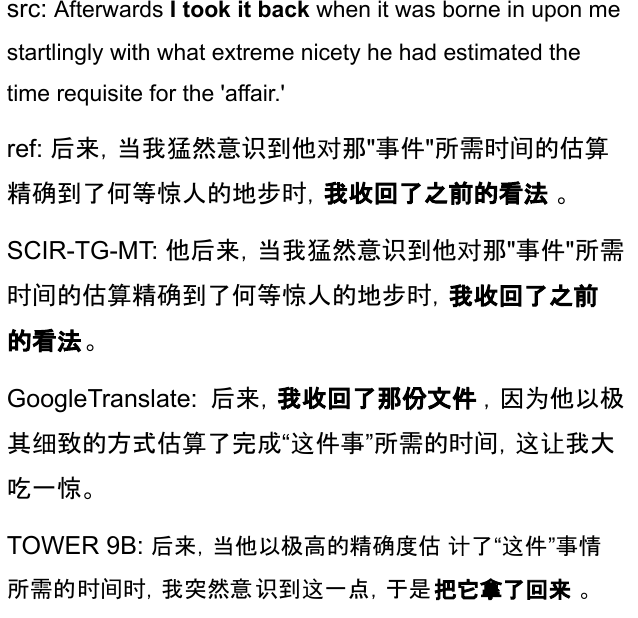}
      
      \vspace{2mm}
      
      \caption{Example on English-Chinese Translation Issues of Figurative Expressions (aa-121).}
      \label{fig:sent-id-121-en-zh}
    }%
  }
\end{figure}

%\begin{figure}[t]
 % \includegraphics[width=.48\textwidth]{Figs/en-zh-example-aa121__cropped.pdf}
  %\caption{Example on English-Chinese Translation Issues of Figurative Expressions (sentence-id: aa-121).}
  %\label{fig:sent-id-121-en-zh}
%\end{figure}

% EN-ZH: 
\textbf{Across-system differences.} The three most apparent error types are Mistranslation, Proofreading, and Terminology, especially for Tower-9B.
The SCIR-TG-MT performed much better than the other two systems. Their total error penalties are (52, 98, 167) respectively for SCIR-TG-MT, GoogleMT, and Tower-9B.
Even though Mistranslation is the common issue of all these three MT systems, SCIR-TG-MT has a much lower overall penalty score 41 than other two systems (79, 113) respectively for GoogleMT and Tower-9B on this error type. 
As one example, the source sentence ``They intrigued and slandered and hated each other only on that account, – but as to effectually \textit{lifting a little finger} – oh, no. By heavens!'' (aa-127) in Figure \ref{fig:sent-id-127-en-zh}, SCIR-TG-MT had the correct translation ``真正动手帮一下忙'' of the sentence and idiom ``(effectually) lifting a little finger'', while the other two systems produced wrong translation (literal) on the idiomatic expression.

Another example can be found via the vMWE ``took it back’’ in the sentence ``Afterwards I \textit{took it back} when it was borne in upon me startlingly with what extreme nicety he had estimated the time requisite for the 'affair'.’’ (sentence-id: aa-121), in Figure \ref{fig:sent-id-121-en-zh}, where SCIR-TG-MT correctly translated it as ``收回了之前的看法 (withdrew my previous opinion)’’ while the other two systems produced ``收回了那份文件  -- took back that file (made up)’’ (GoogleMT) and ``把它拿了回来  -- took it back (literal)’’ (Tower-9B).

\textbf{Across-system similarities.}
Literature text presents greater difficulty than IT document text. This exactly reflects the ``gap'' that LLMs still need to fill in, i.e., the text with rich expressions and idiomatic and metaphorical phrases. For future research, task-specific evaluation frameworks focusing on figurative language can be a growing topic, e.g., grounded metrics on metaphor translation from \newcite{liang2026metahope}, adapted from the general domain HOPE framework.

\subsection{English--German} \label{sec:en-de}

Overall, during annotation we noticed that TRIVE has a considerably more ``literary'' style. It knows and uses German idioms, cf. aa-119 (\textit{Zahn der Zeit}), aa-105 (\textit{kein Auge trocken}), and introduces additional words that lend the text a more literary character (cf. aa-122, aa-115, aa-109). This stylistic richness, however, sometimes comes at the cost of precision.

A translation issue we encountered repeatedly in practice concerns the word ``standard'' (aa-137), which, in an engineering context, should almost exclusively be translated as \textit{Norm}. A \textit{Norm} is a document prescribing how to design, build, test, and document a given thing or process, whereas a standard may also encompass non-binding recommendations. Accordingly, the statement "this does not comply with the standard" (\textit{Dies entspricht nicht dem Standard}) is, in an engineering context, generally less consequential than "this does not meet the Norm" (\textit{Dies erfüllt nicht die Norm}). Failing to meet a standard may impact customer acceptance of a business, whereas failing to meet a \textit{Norm} may entail more severe legal consequences.

\subsection{English--Polish}
% inser Anna's notes here
The English--Polish translations produced by the selected MT systems were of high quality, although performance varied substantially across text genres. Technical texts were translated accurately owing to well-established terminology, whereas literary texts containing idiomatic expressions and MWEs remained considerably more challenging. News-style texts generally exhibited fewer semantic errors because the surrounding context was easier for the systems to infer.

The most frequent error category was stylistic errors (STL), followed by mistranslations (MIS).
Stylistic errors mainly involved unnatural Polish sentence structure, awkward collocations and overly complex wording, while mistranslations often resulted from literal translation of MWEs and idiomatic expressions, sometimes changing the meaning of the entire
sentence.

Examples:
\vspace{-0.3cm}
\begin{itemize}
    \item STL

aa-005: \textit{However, when you apply a filter by selection, the conditional filter already applied on the field is removed before the filter by selection is applied.}

DeepSeek V4 Pro: \textit{Jednakże, gdy zastosujesz filtr według zaznaczenia, warunkowy filtr już zastosowany na tym polu zostanie usunięty przed zastosowaniem filtru według zaznaczenia.}

Comment: The whole sentence is just complicated and akward. It needs to be read several times to be fully understood.

    \item MIS

aa-102: \textit{Shrugging, he gives up and I turn to the \textbf{twice disagreeable chicken} and eat guiltily, my appetite spoiled.}

TRIVE: Wzruszając ramionami, daje za wygraną, a ja wracam do \textbf{podwójnie nieapetycznego kurczaka} i jem z poczuciem winy, z popsutym apetytem.

Comment: This sentence cannot be translated literally while preserving its meaning, and here we are dealing with a literal translation.

\end{itemize}

\textbf{Linguistic observations.}
% This is interesting because it explains why Polish is difficult.
Several errors reflected fundamental linguistic differences between English and Polish: verbal aspect, grammatical gender, singular/plural/formal ``you", idioms, and collocations.

Examples:
\vspace{-0.3cm}
\begin{itemize}
    \item verbal aspect

aa‑012: \textit{When items in a field \textbf{are hidden} by conditional filtering, a funnel appears to the left of the drop-down arrow Field arrow.}

DeepSeek V4 Pro: \textit{Gdy elementy w polu \textbf{są }\textbf{ukrywane }przez filtrowanie warunkowe, po lewej stronie strzałki rozwijanego pola (strzałki pola) pojawia się lejek.}

Comment: It should be a perfective verb \textit{ukryć} used here, instead of imperfective \textit{ukrywać}: \textit{zostały/są ukryte}.

    \item grammatical gender

aa‑124: \textit{\textbf{I heard} the name of Kurtz pronounced, then the words, ``take advantage of this unfortunate accident."}

DeepSeek V4 Pro: \textit{\textbf{Usłyszałem} wymówione imię Kurtza, a potem słowa: ``skorzystajcie z tego nieszczęśliwego wypadku".}

Comment: Polish distinguishes grammatical genders. So it could be here either \textit{Usłyszałem} (masculine), or \textit{Usłyszałam} (feminine). Without a context we cannot be sure whether it is she or he speaking.

    \item singular/plural/formal ``you":
    
    ab‑009:
\textit{Never mind, Mrs. Bray will join\textbf{ you} later.}

DeepSeek V4 Pro: \textit{Nieważne, pani Bray dołączy do\textbf{ was }później.} $\rightarrow$ ``you" in plural.

TRIVE: \textit{Nieważne, pani Bray dołączy do\textbf{ ciebie }później.} $\rightarrow$ ``you" in singular, informal.

Dubformer: \textit{Mniejsza z tym, pani Bray dołączy do \textbf{pana} później. }$\rightarrow$ ``you" in singular, formal (he --  sir).

Comment: All translations are correct. More context is needed to determine which use of ``you" is appropriate here.
\end{itemize}

\textbf{Annotation observations.}
We also noted that sentence-level evaluation without wider
document context occasionally made multiple translations equally
acceptable.
% This is an excellent point.
It shows a limitation of test-suite evaluation rather than MT itself.
This has also been reflected by the literature \cite{gladkoff2026lamppost}.
We list more English--Polish examples of translation issues in Section \ref{sec:additional-en-pl-example}.
% \subsection{Cross-language-pairs?}

\begin{comment}

\textbf{Cross-annotator agreement.}
The English--Polish subset was annotated independently by two native speakers over the same 150 segments and three MT systems (450 MT outputs, 3600 severity cells). The two annotators assigned an identical error profile to 265 of the 450 outputs (58.9\%). Agreement on error \emph{detection} was only fair (Cohen's $\kappa = 0.32$), with 128 outputs flagged by exactly one annotator, and per-segment \textsc{Segs} totals correlate at Pearson $r = 0.70$.
Critically, both annotations yield the \emph{same system ranking}, so the
conclusion of the human evaluation is robust to annotator identity. What differs is resolution rather than order: the more conservative annotator recorded 180 error instances at a mean severity of 1.52 and the stricter one 137 instances at 2.86, which compresses the gap between the best and the worst system from 48 penalty points to 13. We note that this contrasts with the English - Chinese observation in Appendix \ref{sec:additional-en-zh}, where annotators agreed readily on the presence of an error and diverged mainly on its severity. For English--Polish, detection itself was a substantial source of disagreement. Appendix \ref{sec:en-pl-iaa} reports the full comparison in Table \ref{tab:en-pl-iaa} and discusses two structural causes: overlap between error categories, and a difference in the procedure the two annotators followed.

\end{comment}

Two Polish native speakers completed the annotation independently over all 150 segments; inter-annotator agreement is reported in Appendix \ref{sec:en-pl-iaa}. Agreement on error detection was only fair ($\kappa = 0.32$) and severity calibration differed substantially between the two annotators, so the absolute penalty magnitudes reported for English--Polish should be read as annotator-dependent, even though the system ranking is stable across both annotations.
Inter-annotator agreement (IAA) for English--Arabic and English--Chinese is to be reported in the future.
\section{Conclusions and Future Work}
\label{sec:conclusion}

In this work, we used the AlphaMWE multilingual parallel corpus as a
test suite in the WMT 2026 Test Suites Shared Task to investigate
whether multiword expressions and other non-compositional phenomena
remain challenging for contemporary MT systems. We evaluated outputs
from 31 participating systems across English--Chinese,
English--German, English--Polish, Modern Standard Arabic, Egyptian
Arabic, and Tunisian Arabic. We first applied SacreBLEU, chrF++, and
BERTScore F1, ranked systems independently for each language pair,
and selected the top-performing systems through mean rank aggregation.
We then complemented the automatic evaluation with HOPE-based human
evaluation by native speakers.

Across the evaluated language pairs, the human analysis indicates that
high automatic metric performance does not eliminate linguistically
important translation errors. In particular, idioms, verbal MWEs,
metaphorical and figurative expressions, lexical ambiguity, and
language-specific collocations continue to expose weaknesses in
otherwise fluent MT output. These problems are especially visible in
literary and expression-rich text, whereas technical and news-oriented
sentences were generally handled more reliably. The observed errors
also differed across languages: English--Chinese and English--Arabic
showed prominent failures involving literal interpretation of idiomatic
and figurative expressions, English--Polish exposed difficulties
involving verbal aspect, collocation, grammatical distinctions and
context-sensitive lexical choice, while the English--German analysis
highlighted the tension between idiomatic or literary rendering and
domain-specific terminological precision.

The comparison between automatic and human evaluation further suggests
that no single automatic metric provides a complete account of
translation quality. Although the three automatic metrics often agreed
on the strongest systems, non-zero rank standard deviations for several
systems revealed substantial metric disagreement. Human annotation
subsequently identified errors -- particularly those involving semantic
scope, idiomaticity, terminology, and contextual interpretation -- that are difficult to characterize through aggregate automatic scores alone.
This difference is also reflected at the system level: HOPE-based human evaluation does not fully preserve the ranking induced by the aggregated automatic metrics, changing the relative ordering of the selected systems for several language pairs.

\begin{comment}
 
The two language pairs for which we have independent double annotation also show that annotator agreement itself varies considerably by language team
($\kappa = 0.32$ for Polish against $0.10$ for German), while the resulting system rankings remain stable in both cases, this suggests that HOPE-style error taxonomies support reliable system ordering more readily than reliable absolute error accounting.
 
\end{comment}

Our results therefore support combining complementary automatic metrics
with targeted human evaluation when investigating linguistically
challenging translation phenomena.

A further limitation concerns sentence-level evaluation. Several
English source sentences are underspecified when removed from their
wider discourse context. This is particularly important for languages
that require grammatical distinctions not explicitly encoded in
English, such as gender, verbal aspect, or singular/plural and
formal/informal forms of \textit{you}. In such cases, multiple target
translations may be equally valid, and reference-based sentence-level
evaluation can potentially penalize legitimate alternatives. The
AlphaMWE resource retains contextual material originating from the
PARSEME data, which provides an opportunity to address this limitation
in future work through document-level or context-aware translation
evaluation.

Future work will therefore extend the present analysis in three
directions. First, we plan to evaluate context-aware and document-level
translation of MWE-containing sentences. Second, we will investigate
phenomenon-specific evaluation frameworks for idioms, metaphors, and
other figurative expressions, where general-purpose MT metrics may
provide insufficient diagnostic information. Third, larger-scale human
annotation with multiple annotators per language will enable analysis
of inter-annotator agreement, particularly for error severity, and help
separate genuine system errors from variation in linguistic preference
and annotation judgement. Together, these directions can further
develop AlphaMWE from a multilingual test suite into a diagnostic
resource for identifying the remaining linguistic blind spots of
LLM-based machine translation.

% \section{Conclusions and Future Work}

% Context needed, as in  \cite{gladkoff2026lamppost}

% To address sentence-level evaluation issues, the AlphaMWE corpus includes the context story text from the PARSEME dataset \cite{han2026towards,parseme2017shared}, and we will try document-level translation in the future, including such sentences with MWE annotations.

% cross-language pairs common issues/findings: metaphors, idioms, figurative expressions set challenges to LLMs, espacially En-Zh, En-ar, and? 

% Some LLMs had literal expressions as output but at the cost of precision? hallucination?

\section*{Limitations}
The most important limitations are: 1) two annotators for Arabic but only for one round, 75 segments each;
three annotators for Chinese, with two doing 75 segments each, then the 3rd person double-checked the 1st round annotation.
%One Polish native speaker finished one-round annotation, and the 2nd native speaker is still doing the annotation.
%Inter-annotator agreement (IAA) on these 3 language pairs is to be reported in the future.
% English-German had two annotators finish the annotation independently, so the IAA is to be reported for the final report.
Two German native speakers also completed the annotation independently over all 150 segments. Agreement is reported in Appendix \ref{sec:en-de-iaa} and is markedly weaker than for Polish ($\kappa = 0.10$ on error detection), with the two annotators flagging largely disjoint sets of outputs, so the English--German penalty magnitudes should likewise be treated as annotator-dependent.
\begin{comment} 
\end{comment}
% one native annotator per language or limited annotator coverage; 
2) only 150 common segments used for cross-language human comparison.
3) different systems participated in different language pairs.
4) sentence-level context limitations; and 
5) automatic evaluation limited to three metrics, with more metrics considered to be included with extra time, e.g., h/LEPOR \cite{han2013language}, COMET \cite{rei-etal-2020-comet}, and metaphor specific MetaHOPE \cite{liang2026metahope}.

% \section*{Limitations}

% This document does not cover the content requirements for ACL or any
% other specific venue.  Check the author instructions for
% information on
% maximum page lengths, the required ``Limitations'' section,
% and so on.

\section*{Acknowledgments}

LH is grateful to the HumanAI cluster fund from LIACS, Leiden University.
% \textcolor{red}{Do we have any or we need to comment this section?}

% Bibliography entries for the entire Anthology, followed by custom entries
%\bibliography{custom,anthology-overleaf-1,anthology-overleaf-2}

% Custom bibliography entries only
\bibliography{custom}

\appendix
\clearpage
\section{Language Identifiers}
Language identifiers used in the paper are described in Table \ref{tab:language_codes}.

\begin{table*}[t]
\centering
\small
\caption{Language identifiers used throughout this paper and in the WMT~2026 Test Suites Shared Task.}
\label{tab:language_codes}
\begin{tabular}{llll}
\toprule
\textbf{Pair} & \textbf{Code} & \textbf{Language} & \textbf{Description} \\
\midrule
en$\rightarrow$ar  & \texttt{ar}  & Modern Standard Arabic (MSA) & Standard written Arabic \\
en$\rightarrow$arz & \texttt{arz} & Egyptian Arabic              & Egyptian Arabic dialect \\
en$\rightarrow$aeb & \texttt{aeb} & Tunisian Arabic              & Tunisian Arabic dialect \\
en$\rightarrow$de  & \texttt{de}  & German                       & Standard German \\
en$\rightarrow$pl  & \texttt{pl}  & Polish                       & Polish \\
en$\rightarrow$zh  & \texttt{zh}  & Chinese                      & Simplified Chinese \\
\bottomrule
\end{tabular}
\end{table*}

\section{English--Polish Cross-Annotator Agreement}
\label{sec:en-pl-iaa}

Table \ref{tab:en-pl-iaa} summarises the comparison between the two independent
English--Polish annotations. Beyond the aggregate figures, two structural
sources of disagreement are worth recording.

\textbf{Category overlap.} A recurring pattern is that both annotators identify
the \emph{same} defect but file it under different categories. In
\mbox{aa-134}, the untranslated ``Kong'' for Hong Kong was recorded as a proper
name error (PRN) by one annotator and as a missing source adaptation (RAM) by
the other; in \mbox{aa-104}, one annotator recorded terminology (TRM) and the
other mistranslation (MIS) for the same expansion of the source acronym.
Per-category detection $\kappa$ ranges from $-0.02$ (UGR) to $0.40$ (TRM, PRN),
and RAM was never used by one of the two annotators at all. For HOPE-style
schemes we therefore recommend reporting agreement both per category and after
collapsing categories that overlap by design, and supplying worked examples for
the categories that are otherwise left unused.

\textbf{Annotation procedure.} The two annotators worked in different ways. One annotated against a Polish reference translation revised for this study, which was produced by a different person than the annotator. The other did not use the reference column, preferring to translate each source segment independently before scoring. Part of the disagreement reported above therefore reflects a difference in procedure rather than in judgement. The effect is visible in \mbox{aa-104}, where the original reference left the source acronym ``OAS'' untranslated and all three systems expanded it to \textit{OPA} (\textit{Organizacja Pa\'nstw Ameryka\'nskich}), whereas the passage (Amin speaking at Port Louis in July 1976) concerns the Organisation of African Unity. The annotator who worked from her own translation recorded this as a critical terminology error on all three systems, while the reference-based annotation did not flag it. A related factor is that the severity scale was not calibrated between the annotators in advance, so the same scale could be read differently across error types. We therefore recommend that IAA studies of HOPE-style schemes record the annotation procedure and a shared severity key alongside the taxonomy itself.

\begin{comment}

\textbf{Reference version.} 

The two annotators worked against different reference translations: one used a reference revised for this study, the other the original AlphaMWE reference. Several penalties assigned by the second annotator fall on outputs that match the revised reference exactly. In
\mbox{aa-122}, for instance, all three systems were penalised for
\textit{odwracając się plecami do} (``turning one's back on''), which is in fact
the established Polish idiom and the wording adopted in the revised reference.
Conversely, revising the reference exposed an error that neither annotator
caught initially: in \mbox{aa-104} the original reference left the source
acronym ``OAS'' untranslated, and all three systems expanded it to
\textit{OPA} (\textit{Organizacja Państw Amerykańskich}), whereas the passage
(Amin speaking at Port Louis in July 1976) concerns the Organisation of
African Unity. Reference quality is thus not a neutral background condition in
reference-based human evaluation: it shapes what annotators are able to see, and
it should be held constant across annotators when IAA is being measured.

\end{comment}

\begin{table}[t]
\centering
\small
\caption{Cross-annotator comparison for English--Polish over 150 segments
$\times$ 3 MT systems (450 MT outputs). \textsc{Segs} totals are summed error
penalties; lower is better. A1 and A2 denote the two independent annotators.}
\label{tab:en-pl-iaa}
\begin{tabular}{lrr}
\toprule
 & \textbf{A1} & \textbf{A2} \\
\midrule
\multicolumn{3}{l}{\textit{\textsc{Segs} totals}} \\
\quad DeepSeek V4 Pro & 116 & 136 \\
\quad TRIVE           &  68 & 123 \\
\quad Dubformer       &  89 & 133 \\
\quad best--worst spread & 48 & 13 \\
\midrule
\multicolumn{3}{l}{\textit{Annotation profile}} \\
\quad error instances      & 180 & 137 \\
\quad mean severity        & 1.52 & 2.86 \\
\quad instances at 8 or 16 & 2 & 11 \\
\midrule
\multicolumn{3}{l}{\textit{Agreement}} \\
\quad identical error profiles  & \multicolumn{2}{c}{265 / 450 (58.9\%)} \\
\quad cell-level agreement      & \multicolumn{2}{c}{92.9\%} \\
\quad detection $\kappa$        & \multicolumn{2}{c}{0.32} \\
\quad \textsc{Segs} correlation & \multicolumn{2}{c}{$r = 0.70$, $\rho = 0.38$} \\
\quad system ranking            & \multicolumn{2}{c}{identical} \\
\bottomrule
\end{tabular}
\end{table}

\begin{comment}
example text hiden from here     
\end{comment}

\section{Additional English--Polish Translation Examples}
\label{sec:additional-en-pl-example}
% Polish examples for the AlphaMWE English--Polish human evaluation.
%
% Recommended compiler: XeLaTeX or LuaLaTeX.
% If your paper already uses ctex with XeLaTeX/LuaLaTeX, Polish characters
% (ą ć ę ł ń ó ś ź ż) can be entered directly in UTF-8 as below.
%
% Add to the preamble:
% \usepackage[most]{tcolorbox}
%
% Optional compact style used by all examples:
\newtcolorbox{plexample}[1]{
  enhanced,
  breakable,
  colback=gray!3,
  colframe=gray!55,
  boxrule=0.5pt,
  arc=1.5pt,
  left=5pt,
  right=5pt,
  top=4pt,
  bottom=4pt,
  before skip=7pt,
  after skip=7pt,
  fonttitle=\bfseries,
  title={#1}
}

% \subsection{Representative English--Polish Examples}
% \label{sec:pl-qualitative-examples}

The following examples illustrate representative linguistic phenomena
identified during the HOPE-based human evaluation of the
English--Polish test suite. The source sentence, human reference, and
outputs of the three selected systems are shown for each case.
The discussion summarizes the Polish annotator's qualitative analysis.

% ============================================================
\begin{plexample}{Example 1: Literal translation of an idiomatic MWE (MIS) -- aa-042}

\textbf{Source.}
But it did not give me the time of day.

\medskip
\textbf{Reference.}
Ale nie zwrócił na mnie uwagi.

\medskip
\textbf{DeepSeek V4 Pro.}
Ale nie dało mi pory dnia.

\medskip
\textbf{TRIVE.}
Ale nawet nie zaszczyciło mnie spojrzeniem.

\medskip
\textbf{Dubformer.}
Ale on nawet nie raczył na mnie spojrzeć.

\medskip
\textbf{Discussion.}
The DeepSeek V4 Pro output translates the idiom \textit{give me the
time of day} literally as \textit{dało mi pory dnia}, which does not
convey the intended idiomatic meaning in Polish. In contrast, TRIVE
and Dubformer render the expression idiomatically. This example
illustrates a severe MWE-related mistranslation, where lexical
correspondence is preserved but the intended meaning is lost. It is worth noting that the reference translation differs from the correct machine translations in terms of MWE, but conveys the same meaning. 

\end{plexample}

% ============================================================
\begin{plexample}{Example 2: Verbal aspect in Polish (UGR) -- aa-012}

\textbf{Source.}
When items in a field are hidden by conditional filtering, a funnel
appears to the left of the drop-down arrow Field arrow.

\medskip
\textbf{Reference.}
Gdy elementy w polu są ukryte przez filtrowanie warunkowe, na lewo od
strzałki rozwijanej strzałki Pola pojawia się lejek.

\medskip
\textbf{DeepSeek V4 Pro.}
Gdy elementy w polu są ukrywane przez filtrowanie warunkowe, po lewej
stronie strzałki rozwijanego pola (strzałki pola) pojawia się lejek.

\medskip
\textbf{TRIVE.}
Gdy elementy w polu są ukryte przez filtrowanie warunkowe, po lewej
stronie strzałki rozwijania pola pojawia się lejek.

\medskip
\textbf{Dubformer.}
Gdy elementy w polu są ukryte przez filtrowanie warunkowe, na lewo od
strzałki rozwijania (strzałki pola) pojawia się lejek.

\medskip
\textbf{Discussion.}
The DeepSeek V4 Pro translation uses the imperfective form
\textit{ukrywane} (`being hidden repeatedly/in progress'), whereas
the Polish context calls for a resultative or perfective reading,
such as \textit{są ukryte}. This reflects a structural difference
between English and Polish: Polish verbal aspect is strongly
lexicalized and must often be resolved explicitly during translation.
The error does not prevent comprehension, but it produces a less
natural grammatical interpretation.

\end{plexample}

% ============================================================
\begin{plexample}{Example 3: Multiple acceptable context-dependent translations -- aa-044}

\textbf{Source.}
Uncle Vernon might still have been able to make his deal -- if it
hadn't been for the owl.

\medskip
\textbf{Reference.}
Wujek Vernon nadal mógłby zawrzeć umowę -- gdyby nie sowa.

\medskip
\textbf{DeepSeek V4 Pro.}
Wuj Vernon mógłby jeszcze dopiąć swego -- gdyby nie ta sowa.

\medskip
\textbf{TRIVE.}
Wuj Vernon być może zdołałby jeszcze ubić interes -- gdyby nie sowa.

\medskip
\textbf{Dubformer.}
Wuj Vernon być może wciąż byłby w stanie dobić targu -- gdyby nie
sowa.

\medskip
\textbf{Discussion.}
All three system outputs are plausible Polish translations, but
\textit{dopiąć swego}, \textit{ubić interes}, and \textit{dobić
targu} introduce slightly different semantics.
The example illustrates a limitation of sentence-level evaluation:
without broader discourse context, more than one target-language
realization may be acceptable, and the reference should not be
treated as the only valid formulation.

\end{plexample}

% ============================================================
\begin{plexample}{Example 4: Collocational naturalness (STL) -- aa-106}

\textbf{Source.}
We step into the street and my friend David Shahar, whose chest is
large, takes a deep breath and advises me to do the same.

\medskip
\textbf{Reference.}
Wchodzimy na ulicę i mój przyjaciel David Shahar, którego klatka
piersiowa jest duża, bierze głęboki oddech i radzi mi zrobić to samo.

\medskip
\textbf{DeepSeek V4 Pro.}
Wychodzimy na ulicę, a mój przyjaciel David Shahar, którego klatka
piersiowa jest duża, bierze głęboki oddech i radzi mi zrobić to samo.

\medskip
\textbf{TRIVE.}
Wychodzimy na ulicę i mój przyjaciel David Shahar, który ma szeroką
klatkę piersiową, bierze głęboki oddech i radzi mi zrobić to samo.

\medskip
\textbf{Dubformer.}
Wychodzimy na ulicę, a mój przyjaciel David Shahar, który ma postawną
klatkę piersiową, bierze głęboki oddech i radzi mi zrobić to samo.

\medskip
\textbf{Discussion.}
Dubformer uses \textit{postawną klatkę piersiową}. In Polish,
\textit{postawny} is normally used to describe a person's build
rather than a body part, making the phrase collocationally unnatural.
The translation remains understandable, so the issue is primarily
stylistic rather than a major semantic failure.

\end{plexample}

% ============================================================
\begin{plexample}{Example 5: Domain terminology versus literal wording (TRM) -- aa-137}

\textbf{Source.}
The fiscal advantages for clean vehicles could then be introduced
some two years before the new standards take effect.

\medskip
\textbf{Reference.}
Korzyści podatkowe dla ekologicznie czystych pojazdów mogłyby zostać
wprowadzone na około dwa lata przed wejściem w życie nowych norm.

\medskip
\textbf{DeepSeek V4 Pro.}
Ulgi fiskalne dla pojazdów ekologicznych mogłyby wówczas zostać
wprowadzone na około dwa lata przed wejściem w życie nowych norm.

\medskip
\textbf{TRIVE.}
Zachęty fiskalne dla czystych pojazdów mogłyby wówczas zostać
wprowadzone na około dwa lata przed wejściem w życie nowych norm.

\medskip
\textbf{Dubformer.}
Korzyści fiskalne dla czystych pojazdów mogłyby wówczas zostać
wprowadzone na około dwa lata przed wejściem w życie nowych norm.

\medskip
\textbf{Discussion.}
The literal phrase \textit{czyste pojazdy} is understandable, but
Polish domain usage normally favors established expressions such as
\textit{pojazdy ekologiczne} or \textit{pojazdy niskoemisyjne}.
DeepSeek V4 Pro therefore provides the most idiomatic terminology in
this example, while the other outputs illustrate how literal lexical
transfer can reduce terminological adequacy even when the overall
meaning remains recoverable.

\end{plexample}

% ============================================================
\begin{plexample}{Example 6: Literal target-language construction (STL) -- aa-133}

\textbf{Source.}
Greece, Italy, France, Spain and Austria, have understood that
something has to be done and now have a presence in Albania.

\medskip
\textbf{Reference.}
Grecja, Włochy, Francja, Hiszpania i Austria zrozumiały, że trzeba
coś zrobić i teraz są obecne w Albanii.

\medskip
\textbf{DeepSeek V4 Pro.}
Grecja, Włochy, Francja, Hiszpania i Austria zrozumiały, że trzeba
coś zrobić, i teraz mają swoją obecność w Albanii.

\medskip
\textbf{TRIVE.}
Grecja, Włochy, Francja, Hiszpania i Austria zrozumiały, że coś
trzeba zrobić, i są teraz obecne w Albanii.

\medskip
\textbf{Dubformer.}
Grecja, Włochy, Francja, Hiszpania i Austria zrozumiały, że trzeba
coś zrobić i są teraz obecne w Albanii.

\medskip
\textbf{Discussion.}
DeepSeek V4 Pro transfers the English construction \textit{have a
presence} too directly as \textit{mają swoją obecność}. Although the
meaning is transparent, this construction is not idiomatic Polish.
The reference, TRIVE, and Dubformer instead use \textit{są obecne},
which is considerably more natural. The example demonstrates how
literal translation can produce stylistic errors without necessarily
causing a loss of propositional meaning.

\end{plexample}

% ============================================================
\begin{plexample}{Example 7: Scope ambiguity in MWE translation (MIS/STL) -- aa-108}

\textbf{Source.}
In 1947 Copeland had been sent to Damascus ("by whom is not stated,"
Kedourie says) "to make unofficial contact" with Syrian leaders and
"to probe for means of persuading them, on their own, to liberalize
their political system."

\medskip
\textbf{Reference.}
W 1947 r. Copeland został wysłany do Damaszku ("przez kogo nie jest
powiedziane", mówi Kedourie), aby "nawiązać nieoficjalny kontakt" z
syryjskimi przywódcami i "poszukać sposobów przekonania ich, na
własną rękę, do liberalizacji ich systemu politycznego".

\medskip
\textbf{DeepSeek V4 Pro.}
W 1947 roku Copeland został wysłany do Damaszku („przez kogo, nie
podano”, mówi Kedourie) „aby nawiązać nieoficjalny kontakt” z
syryjskimi przywódcami oraz „aby zbadać sposoby nakłonienia ich,
samodzielnie, do liberalizacji ich systemu politycznego”.

\medskip
\textbf{TRIVE.}
W 1947 roku Copeland został wysłany do Damaszku („nie podano przez
kogo”, jak twierdzi Kedourie), „aby nawiązać nieoficjalny kontakt” z
syryjskimi przywódcami i „wybadać sposoby nakłonienia ich, by z
własnej inicjatywy zliberalizowali swój system polityczny”.

\medskip
\textbf{Dubformer.}
W 1947 roku Copeland został wysłany do Damaszku („przez kogo, nie
podano”, pisze Kedourie) „w celu nawiązania nieoficjalnych
kontaktów” z syryjskimi przywódcami i „zbadania możliwości
przekonania ich, aby sami zliberalizowali swój system polityczny”.

\medskip
\textbf{Discussion.}
The placement of DeepSeek V4 Pro's \textit{samodzielnie} makes the
scope of \textit{on their own} difficult to recover and can suggest
an unintended interpretation. TRIVE and Dubformer make the semantic
relation explicit by using \textit{z własnej inicjatywy} and
\textit{aby sami zliberalizowali}, respectively. This case shows
that an MWE or modifier may be lexically translated while its
syntactic scope is altered, potentially changing the interpretation
of the whole sentence.

\end{plexample}

% Cross annotator agreement German
\section{English--German Cross-Annotator Agreement}
\label{sec:en-de-iaa}
 
Table \ref{tab:en-de-iaa} summarises the comparison between the two independent
English--German annotations, computed exactly as for English--Polish.
 
\textbf{Unused and disjointly used categories.} Annotator A2 did not use the
terminology (TRM) or proper-name (PRN) categories at any point, so the
terminological analysis reported in Section \ref{sec:en-de} (including the
\textit{standard}/\textit{Norm} distinction in \mbox{aa-137}) has no counterpart in the second annotation. More strikingly, both annotators made heavy use of the impact category, 87 times between them, yet applied it to the same MT output on only three occasions (per-category $\kappa = -0.04$); a further 63 style penalties appear in A2's annotation alone. The only category with substantial agreement is required-adaptation-missing (RAM, $\kappa = 0.57$), which both annotators applied to the uncorrected source defect in \mbox{aa-134}, where the source reads ``Kong's civil service'' for Hong Kong. These figures suggest that the divergence is driven less by differing severity thresholds than by differing readings of what the categories denote, and they reinforce the recommendation made for English--Polish in Appendix \ref{sec:en-pl-iaa}: HOPE-style schemes need worked examples per category, and agreement should be reported per category rather than only in aggregate.
 
\textbf{Effect on discrimination.} The two annotations separate the systems by different margins. Under A1, TRIVE and DeepSeek V4 Pro differ by 5 penalty points out of roughly 140, which is well within annotation noise, while A2 separates them by 34. Conversely, A1 places VoxNexus\_V1 far ahead of both competitors (42 against 140 and 145), whereas A2's margin is narrower (77 against 97 and 131). The ordering is preserved in both cases, but the confidence one can place in the second-versus-third comparison depends on which annotation is consulted.
 
\begin{table}[t]
\centering
\small
\caption{Cross-annotator comparison for English--German over 150 segments
$\times$ 3 MT systems (450 MT outputs). \textsc{Segs} totals are summed error
penalties; lower is better. A1 and A2 denote the two independent annotators.}
\label{tab:en-de-iaa}
\begin{tabular}{lrr}
\toprule
 & \textbf{A1} & \textbf{A2} \\
\midrule
\multicolumn{3}{l}{\textit{\textsc{Segs} totals}} \\
\quad VoxNexus\_V1     &  42 &  77 \\
\quad DeepSeek V4 Pro  & 145 & 131 \\
\quad TRIVE            & 140 &  97 \\
\quad best--worst spread & 103 & 54 \\
\midrule
\multicolumn{3}{l}{\textit{Annotation profile}} \\
\quad error instances      & 126 & 162 \\
\quad mean severity        & 2.60 & 1.88 \\
\quad instances at 8 or 16 & 12 & 3 \\
\midrule
\multicolumn{3}{l}{\textit{Agreement}} \\
\quad identical error profiles  & \multicolumn{2}{c}{252 / 450 (56.0\%)} \\
\quad cell-level agreement      & \multicolumn{2}{c}{92.7\%} \\
\quad both found an error       & \multicolumn{2}{c}{38} \\
\quad flagged by one only       & \multicolumn{2}{c}{162 (49 A1, 113 A2)} \\
\quad detection $\kappa$        & \multicolumn{2}{c}{0.10} \\
\quad \textsc{Segs} correlation & \multicolumn{2}{c}{$r = 0.15$, $\rho = 0.12$} \\
\quad system ranking            & \multicolumn{2}{c}{identical} \\
\bottomrule
\end{tabular}
\end{table}

\begin{comment}
\end{comment}

\section{Additional English--Chinese Discussion}
\label{sec:additional-en-zh}

The \textbf{annotation tasks require domain-expertise knowledge} for better understanding of the terminologies and expressions. 
As one example, from  sent-id: aa-76: 
\begin{plexample}

    \textbf{Source.} A query that uses wildcard characters in a criteria expression can produce different results under each query mode.
    
    \textbf{Reference.} 在标准表达式中使用通配符的查询，在不同的查询模式下会产生不同的结果。
    
    \textbf{Tower-9B.} 在每个查询模式下，使用通配符的条件表达式可能 会产生不同的结果。
\end{plexample}
It is assigned 8 (major error) by our annotator (literature and translation studies background), for MIS error. However, it actually carried a similar meaning to the reference translation, but they are in very different word orders, which might have an impact to our annotators who are using/looking at the offered reference from the original AlphaMWE corpus.
After discussion, we changed it to unnatural/proof-reading error with severity level 2.

\textbf{Error-severity agreements.}
It is not an issue to agree on ``whether a sentence has an error’’, but a problem on the error severity levels, similar findings from \newcite{liang2026metahope}. E.g., the sentence ``Harry just caught sight of a pair of bright brown eyes staring at him before it closed with a snap.’’ (Sent-id: 054). We have:
\begin{plexample}
    
    \textbf{Reference.} 哈利刚看到一双明亮的棕色眼睛盯着他，然后就啪的一声闭上了。
    
    \textbf{SCIR-TG.}哈利只来得及瞥见一双明亮的棕色眼睛正盯着他，然后门就啪的一声关上了。
    
    \textbf{GoogleMT.} (v) 哈利瞥见一双明亮的棕色眼睛正盯着他，然后那双眼睛啪嗒一声闭上了。
    
    \textbf{Tower-9B.} 哈利刚瞥见一对明亮的棕色眼睛在盯着他，然后门就砰地一声关上了。

\end{plexample}

Both SCIR-TG and Tower-9B translated ``door closed’’ while GoogleMT made a correct translation of ``eyes closed’’.  Our initial annotation gave a score of 2 for medium-level; however, after discussion, we think it should be ``4 or 8’’ for major or severe errors.
% => 4? For Min Ma’s 

% => en-zh; we had two-round, i.e. double-check/ cross-checking of the first round annotation.

\section{Detailed Automatic Scores Per System Per Language Pair}
\label{sec:auto-scores-sys-langs}
The detailed scores from SacreBLEU, ChrF++, and BERTscore on submitted systems per language pair are listed in Table \ref{tab:alphamwe_sacrebleu},
\ref{tab:alphamwe_chrfpp}, and 
\ref{tab:alphamwe_bertscore}.

% Required packages:
% \usepackage{booktabs}
% \usepackage{graphicx}

\begin{table*}[t]
\centering
% \small
% \scriptsize 
\tiny 
\caption{SacreBLEU scores for the submitted AlphaMWE test suites. Higher scores are better.}
\label{tab:alphamwe_sacrebleu}
\resizebox{\textwidth}{!}{%
\begin{tabular}{lrrrrrr}
\toprule
\textbf{System} & \textbf{en$\rightarrow$aeb} & \textbf{en$\rightarrow$ar} & \textbf{en$\rightarrow$arz} & \textbf{en$\rightarrow$de} & \textbf{en$\rightarrow$pl} & \textbf{en$\rightarrow$zh} \\
\midrule
CUNI-AR & 8.46 & 30.02 & 9.03 & -- & -- & -- \\
CUNI-EdUKate & -- & -- & -- & 40.52 & -- & -- \\
CUNI-MH-v3 & -- & -- & -- & 41.38 & -- & -- \\
Cohere CAT+ & 3.04 & 23.97 & 3.89 & 39.16 & 29.79 & 32.88 \\
Command A+ & 2.69 & 21.88 & 2.35 & 36.03 & 26.89 & 40.04 \\
DeepSeek V4 Pro & 7.35 & 28.75 & 9.90 & 50.00 & \textbf{40.69} & 37.94 \\
Dubformer & 7.01 & 22.80 & 8.72 & 48.30 & 40.52 & 43.89 \\
GLM 4 - 9B & 0.91 & 17.42 & 2.22 & 36.80 & 31.41 & 43.32 \\
GPT 5.5 & 6.95 & 26.84 & \textbf{10.32} & 48.37 & 38.20 & 42.81 \\
GPT OSS 120B & 3.74 & 23.14 & 6.71 & 40.08 & 34.11 & 39.53 \\
GPT OSS 20B & 4.19 & 19.21 & 4.71 & 31.24 & 27.43 & 35.14 \\
Gemini 3.1 Pro & 1.05 & 5.48 & 1.31 & 7.09 & 6.57 & 7.35 \\
Gemma 4 - 31B & 5.94 & 27.73 & 9.83 & 47.97 & 36.67 & 42.16 \\
Gemma 4 - 4B & 7.01 & 31.64 & 2.72 & 44.56 & 37.42 & 44.23 \\
GoogleTranslate & 7.79 & \textbf{36.46} & 2.42 & 47.68 & 38.59 & \textbf{47.23} \\
HW-TSC & -- & 26.84 & -- & -- & -- & 35.11 \\
Lumen & -- & 28.26 & 8.84 & 46.77 & 36.83 & 40.37 \\
Ministral 3 - 14B & 2.05 & 21.61 & 2.08 & 32.74 & 29.14 & 37.62 \\
Mistral Medium 3.5 & 6.62 & 28.70 & 4.87 & 44.67 & 36.83 & 43.24 \\
Qingqiu-MT-9B & 6.41 & 20.76 & 8.86 & 46.31 & 28.10 & 40.33 \\
Qwen 3.5 - 9b & 3.90 & 19.13 & 1.68 & 31.58 & 22.99 & 30.87 \\
Qwen 3.6 - 27b & 4.38 & 22.08 & 4.01 & 38.52 & 27.76 & 36.94 \\
SCIR-TG-MT & -- & -- & -- & -- & -- & 45.48 \\
SRPOL & -- & -- & 4.11 & 39.49 & -- & 35.90 \\
SalamandraTA & 6.13 & 22.63 & 5.03 & 46.54 & 31.05 & 43.06 \\
TRIVE & \textbf{8.66} & 30.92 & 9.12 & 49.01 & 40.51 & 40.34 \\
Tiny Aya Global & -- & 21.88 & 0.07 & 29.25 & 20.55 & 40.55 \\
Tower 9B & 0.73 & 8.85 & 0.53 & 44.67 & 33.05 & 44.97 \\
UvA-MT & -- & -- & 9.83 & 43.70 & -- & 36.11 \\
VoxNexus\_V1 & -- & 23.15 & 9.41 & \textbf{50.04} & -- & -- \\
Wayfinder & -- & 26.52 & 10.27 & 47.43 & 37.44 & 39.63 \\
\bottomrule
\end{tabular}%
}
\end{table*}

\begin{table*}[t]
\centering
% \small
\tiny 

\caption{chrF++ scores for the submitted AlphaMWE test suites. Higher scores are better.}
\label{tab:alphamwe_chrfpp}
\resizebox{\textwidth}{!}{%
\begin{tabular}{lrrrrrr}
\toprule
\textbf{System} & \textbf{en$\rightarrow$aeb} & \textbf{en$\rightarrow$ar} & \textbf{en$\rightarrow$arz} & \textbf{en$\rightarrow$de} & \textbf{en$\rightarrow$pl} & \textbf{en$\rightarrow$zh} \\
\midrule
CUNI-AR & 34.50 & 59.72 & 39.58 & -- & -- & -- \\
CUNI-EdUKate & -- & -- & -- & 63.07 & -- & -- \\
CUNI-MH-v3 & -- & -- & -- & 63.41 & -- & -- \\
Cohere CAT+ & 13.75 & 54.41 & 28.32 & 62.97 & 56.02 & 23.80 \\
Command A+ & 17.68 & 52.65 & 23.78 & 60.54 & 53.47 & 27.17 \\
DeepSeek V4 Pro & 32.76 & 58.67 & 39.36 & 69.83 & \textbf{63.40} & 27.93 \\
Dubformer & 31.07 & 54.43 & 38.05 & 68.59 & 62.78 & 30.60 \\
GLM 4 - 9B & 0.85 & 47.84 & 22.94 & 60.60 & 55.24 & 29.19 \\
GPT 5.5 & 33.63 & 57.29 & 39.98 & 69.13 & 61.84 & 29.96 \\
GPT OSS 120B & 26.59 & 54.23 & 33.56 & 63.33 & 58.28 & 26.35 \\
GPT OSS 20B & 26.02 & 48.94 & 30.34 & 56.88 & 52.80 & 23.96 \\
Gemini 3.1 Pro & 16.90 & 35.85 & 19.29 & 38.70 & 37.24 & 12.67 \\
Gemma 4 - 31B & 27.51 & 58.37 & 39.84 & 68.85 & 60.16 & 29.64 \\
Gemma 4 - 4B & 30.36 & 61.40 & 24.77 & 66.44 & 60.26 & 29.07 \\
GoogleTranslate & 31.60 & \textbf{65.24} & 24.92 & 68.62 & 60.97 & \textbf{32.16} \\
HW-TSC & -- & 57.19 & -- & -- & -- & 24.76 \\
Lumen & -- & 57.68 & 38.69 & 67.56 & 60.38 & 28.06 \\
Ministral 3 - 14B & 6.11 & 50.15 & 22.15 & 59.97 & 55.71 & 25.35 \\
Mistral Medium 3.5 & 28.04 & 57.86 & 31.06 & 66.28 & 60.19 & 28.73 \\
Qingqiu-MT-9B & 30.11 & 49.63 & 39.90 & 67.56 & 53.36 & 28.51 \\
Qwen 3.5 - 9b & 23.61 & 49.21 & 23.11 & 57.57 & 50.50 & 21.76 \\
Qwen 3.6 - 27b & 27.23 & 52.42 & 29.01 & 62.59 & 54.08 & 25.92 \\
SCIR-TG-MT & -- & -- & -- & -- & -- & 30.89 \\
SRPOL & -- & -- & 28.29 & 63.63 & -- & 26.28 \\
SalamandraTA & 28.99 & 50.44 & 29.83 & 67.41 & 54.05 & 28.59 \\
TRIVE & \textbf{36.84} & 60.35 & 39.97 & 69.43 & 62.89 & 28.28 \\
Tiny Aya Global & -- & 52.43 & 0.41 & 57.76 & 54.02 & 28.16 \\
Tower 9B & 0.74 & 30.27 & 9.67 & 66.66 & 57.80 & 30.51 \\
UvA-MT & -- & -- & 38.62 & 65.41 & -- & 25.73 \\
VoxNexus\_V1 & -- & 54.35 & 39.14 & \textbf{69.92} & -- & -- \\
Wayfinder & -- & 56.74 & \textbf{40.33} & 68.13 & 60.85 & 27.50 \\
\bottomrule
\end{tabular}%
}
\end{table*}

\begin{table*}[t]
\centering
% \small
\tiny 
\caption{BERTScore F1 scores for the submitted AlphaMWE test suites. Higher scores are better.}
\label{tab:alphamwe_bertscore}
\resizebox{\textwidth}{!}{%
\begin{tabular}{lrrrrrr}
\toprule
\textbf{System} & \textbf{en$\rightarrow$aeb} & \textbf{en$\rightarrow$ar} & \textbf{en$\rightarrow$arz} & \textbf{en$\rightarrow$de} & \textbf{en$\rightarrow$pl} & \textbf{en$\rightarrow$zh} \\
\midrule
CUNI-AR & 0.7779 & 0.8935 & 0.7868 & -- & -- & -- \\
CUNI-EdUKate & -- & -- & -- & 0.8877 & -- & -- \\
CUNI-MH-v3 & -- & -- & -- & 0.8906 & -- & -- \\
Cohere CAT+ & 0.6850 & 0.8756 & 0.7349 & 0.8867 & 0.8598 & 0.8382 \\
Command A+ & 0.6784 & 0.8683 & 0.7049 & 0.8757 & 0.8561 & 0.8529 \\
DeepSeek V4 Pro & 0.7624 & 0.8914 & 0.7810 & 0.9061 & \textbf{0.8906} & 0.8587 \\
Dubformer & 0.7590 & 0.8769 & 0.7745 & 0.9053 & 0.8872 & 0.8668 \\
GLM 4 - 9B & 0.6414 & 0.8579 & 0.7056 & 0.8799 & 0.8623 & 0.8623 \\
GPT 5.5 & 0.7705 & 0.8852 & 0.7790 & 0.9036 & 0.8824 & 0.8632 \\
GPT OSS 120B & 0.7303 & 0.8721 & 0.7565 & 0.8841 & 0.8672 & 0.8495 \\
GPT OSS 20B & 0.7181 & 0.8476 & 0.7439 & 0.8587 & 0.8486 & 0.8381 \\
Gemini 3.1 Pro & 0.5947 & 0.7077 & 0.6181 & 0.6734 & 0.6927 & 0.6726 \\
Gemma 4 - 31B & 0.7474 & 0.8894 & \textbf{0.7883} & 0.9055 & 0.8780 & 0.8613 \\
Gemma 4 - 4B & 0.7423 & 0.8968 & 0.7141 & 0.8993 & 0.8800 & 0.8634 \\
GoogleTranslate & 0.7366 & \textbf{0.9070} & 0.7109 & 0.9013 & 0.8809 & 0.8675 \\
HW-TSC & -- & 0.8819 & -- & -- & -- & 0.8330 \\
Lumen & -- & 0.8857 & 0.7804 & 0.9009 & 0.8776 & 0.8543 \\
Ministral 3 - 14B & 0.6332 & 0.8460 & 0.6886 & 0.8731 & 0.8629 & 0.8453 \\
Mistral Medium 3.5 & 0.7437 & 0.8873 & 0.7567 & 0.8976 & 0.8816 & 0.8630 \\
Qingqiu-MT-9B & 0.7543 & 0.8562 & 0.7845 & 0.9014 & 0.8548 & 0.8557 \\
Qwen 3.5 - 9b & 0.7148 & 0.8612 & 0.7069 & 0.8701 & 0.8444 & 0.8305 \\
Qwen 3.6 - 27b & 0.7242 & 0.8665 & 0.7349 & 0.8857 & 0.8546 & 0.8506 \\
SCIR-TG-MT & -- & -- & -- & -- & -- & \textbf{0.8681} \\
SRPOL & -- & -- & 0.7307 & 0.8874 & -- & 0.8492 \\
SalamandraTA & 0.7460 & 0.8506 & 0.7507 & 0.9010 & 0.8613 & 0.8618 \\
TRIVE & \textbf{0.7857} & 0.8954 & 0.7840 & 0.9056 & 0.8887 & 0.8559 \\
Tiny Aya Global & -- & 0.8645 & 0.5926 & 0.8825 & 0.8664 & 0.8497 \\
Tower 9B & 0.6379 & 0.7914 & 0.6440 & 0.8989 & 0.8685 & 0.8678 \\
UvA-MT & -- & -- & 0.7766 & 0.8944 & -- & 0.8432 \\
VoxNexus\_V1 & -- & 0.8760 & 0.7807 & \textbf{0.9082} & -- & -- \\
Wayfinder & -- & 0.8830 & 0.7850 & 0.9021 & 0.8803 & 0.8554 \\
\bottomrule
\end{tabular}%
}
\end{table*}

 % insert this file content? %  'alpha_mwe_scores_with_per_language_ranks.csv'

% or this file? 
% alpha_mwe_evaluation_results.xlsx
Their corresponding rankings are displayed in Table \ref{tab:rank-sacrebleu}, \ref{tab:rank-chrfpp}, and \ref{tab:rank-bertscore}.

\begin{figure*}[t]
  \includegraphics[width=.99\textwidth]{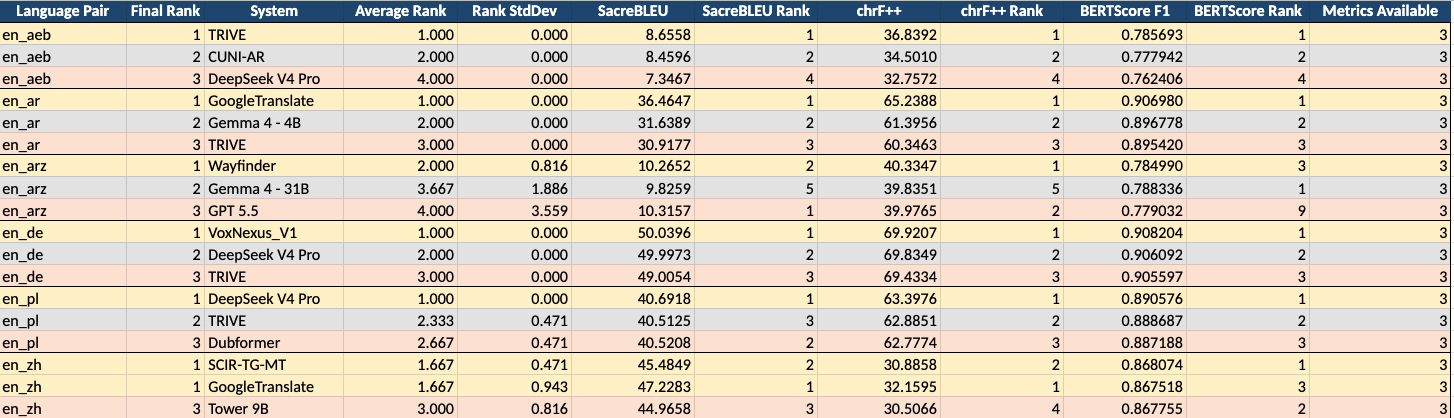}
  \caption{Top 3 MT systems selected for human evaluation: ranking and original scores}
  \label{fig:top3per-langs}
\end{figure*}

The top 3 systems per language pair are listed in Figure \ref{fig:top3per-langs} with their corresponding original metric scores.

% Required packages:
% \usepackage{booktabs}
% \usepackage{graphicx}

\begin{table*}[t]
\centering
% \small
\tiny 
\caption{System rankings based on SacreBLEU for the AlphaMWE test suites. Rank 1 denotes the best-performing system.}
\label{tab:rank-sacrebleu}
\resizebox{\textwidth}{!}{%
\begin{tabular}{lrrrrrr}
\toprule
\textbf{System} & \textbf{en$\rightarrow$aeb} & \textbf{en$\rightarrow$ar} & \textbf{en$\rightarrow$arz} & \textbf{en$\rightarrow$de} & \textbf{en$\rightarrow$pl} & \textbf{en$\rightarrow$zh} \\
\midrule
CUNI-AR & 2 & 4 & 8 & -- & -- & -- \\
CUNI-EdUKate & -- & -- & -- & 17 & -- & -- \\
CUNI-MH-v3 & -- & -- & -- & 16 & -- & -- \\
Cohere CAT+ & 16 & 12 & 18 & 20 & 15 & 25 \\
Command A+ & 17 & 19 & 21 & 23 & 20 & 15 \\
DeepSeek V4 Pro & 4 & 5 & 3 & 2 & \textbf{1} & 18 \\
Dubformer & 6 & 15 & 11 & 5 & 2 & 5 \\
GLM 4 - 9B & 20 & 24 & 22 & 22 & 13 & 6 \\
GPT 5.5 & 7 & 9 & \textbf{1} & 4 & 5 & 9 \\
GPT OSS 120B & 15 & 14 & 12 & 18 & 11 & 17 \\
GPT OSS 20B & 13 & 22 & 15 & 26 & 19 & 23 \\
Gemini 3.1 Pro & 19 & 26 & 25 & 28 & 23 & 27 \\
Gemma 4 - 31B & 11 & 8 & 5 & 6 & 10 & 10 \\
Gemma 4 - 4B & 5 & 2 & 19 & 14 & 7 & 4 \\
GoogleTranslate & 3 & \textbf{1} & 20 & 7 & 4 & \textbf{1} \\
HW-TSC & -- & 10 & -- & -- & -- & 24 \\
Lumen & -- & 7 & 10 & 9 & 8 & 12 \\
Ministral 3 - 14B & 18 & 20 & 23 & 24 & 16 & 19 \\
Mistral Medium 3.5 & 8 & 6 & 14 & 12 & 9 & 7 \\
Qingqiu-MT-9B & 9 & 21 & 9 & 11 & 17 & 14 \\
Qwen 3.5 - 9b & 14 & 23 & 24 & 25 & 21 & 26 \\
Qwen 3.6 - 27b & 12 & 17 & 17 & 21 & 18 & 20 \\
SCIR-TG-MT & -- & -- & -- & -- & -- & 2 \\
SRPOL & -- & -- & 16 & 19 & -- & 22 \\
SalamandraTA & 10 & 16 & 13 & 10 & 14 & 8 \\
TRIVE & \textbf{1} & 3 & 7 & 3 & 3 & 13 \\
Tiny Aya Global & -- & 18 & 27 & 27 & 22 & 11 \\
Tower 9B & 21 & 25 & 26 & 13 & 12 & 3 \\
UvA-MT & -- & -- & 4 & 15 & -- & 21 \\
VoxNexus\_V1 & -- & 13 & 6 & \textbf{1} & -- & -- \\
Wayfinder & -- & 11 & 2 & 8 & 6 & 16 \\
\bottomrule
\end{tabular}%
}
\end{table*}

\begin{table*}[t]
\centering
% \small
\tiny 
\caption{System rankings based on chrF++ for the AlphaMWE test suites. Rank 1 denotes the best-performing system.}
\label{tab:rank-chrfpp}
\resizebox{\textwidth}{!}{%
\begin{tabular}{lrrrrrr}
\toprule
\textbf{System} & \textbf{en$\rightarrow$aeb} & \textbf{en$\rightarrow$ar} & \textbf{en$\rightarrow$arz} & \textbf{en$\rightarrow$de} & \textbf{en$\rightarrow$pl} & \textbf{en$\rightarrow$zh} \\
\midrule
CUNI-AR & 2 & 4 & 6 & -- & -- & -- \\
CUNI-EdUKate & -- & -- & -- & 19 & -- & -- \\
CUNI-MH-v3 & -- & -- & -- & 17 & -- & -- \\
Cohere CAT+ & 18 & 13 & 17 & 20 & 13 & 25 \\
Command A+ & 16 & 16 & 21 & 23 & 19 & 17 \\
DeepSeek V4 Pro & 4 & 5 & 7 & 2 & \textbf{1} & 15 \\
Dubformer & 6 & 12 & 11 & 7 & 3 & 3 \\
GLM 4 - 9B & 20 & 24 & 23 & 22 & 15 & 7 \\
GPT 5.5 & 3 & 9 & 2 & 4 & 4 & 5 \\
GPT OSS 120B & 13 & 15 & 12 & 18 & 11 & 18 \\
GPT OSS 20B & 14 & 23 & 14 & 27 & 21 & 24 \\
Gemini 3.1 Pro & 17 & 25 & 25 & 28 & 23 & 27 \\
Gemma 4 - 31B & 11 & 6 & 5 & 5 & 10 & 6 \\
Gemma 4 - 4B & 7 & 2 & 20 & 13 & 8 & 8 \\
GoogleTranslate & 5 & \textbf{1} & 19 & 6 & 5 & \textbf{1} \\
HW-TSC & -- & 10 & -- & -- & -- & 23 \\
Lumen & -- & 8 & 9 & 10 & 7 & 14 \\
Ministral 3 - 14B & 19 & 20 & 24 & 24 & 14 & 22 \\
Mistral Medium 3.5 & 10 & 7 & 13 & 14 & 9 & 9 \\
Qingqiu-MT-9B & 8 & 21 & 4 & 9 & 20 & 11 \\
Qwen 3.5 - 9b & 15 & 22 & 22 & 26 & 22 & 26 \\
Qwen 3.6 - 27b & 12 & 18 & 16 & 21 & 16 & 20 \\
SCIR-TG-MT & -- & -- & -- & -- & -- & 2 \\
SRPOL & -- & -- & 18 & 16 & -- & 19 \\
SalamandraTA & 9 & 19 & 15 & 11 & 17 & 10 \\
TRIVE & \textbf{1} & 3 & 3 & 3 & 2 & 12 \\
Tiny Aya Global & -- & 17 & 27 & 25 & 18 & 13 \\
Tower 9B & 21 & 26 & 26 & 12 & 12 & 4 \\
UvA-MT & -- & -- & 10 & 15 & -- & 21 \\
VoxNexus\_V1 & -- & 14 & 8 & \textbf{1} & -- & -- \\
Wayfinder & -- & 11 & \textbf{1} & 8 & 6 & 16 \\
\bottomrule
\end{tabular}%
}
\end{table*}

\begin{table*}[t]
\centering
% \small
\tiny 
\caption{System rankings based on BERTScore F1 for the AlphaMWE test suites. Rank 1 denotes the best-performing system.}
\label{tab:rank-bertscore}
\resizebox{\textwidth}{!}{%
\begin{tabular}{lrrrrrr}
\toprule
\textbf{System} & \textbf{en$\rightarrow$aeb} & \textbf{en$\rightarrow$ar} & \textbf{en$\rightarrow$arz} & \textbf{en$\rightarrow$de} & \textbf{en$\rightarrow$pl} & \textbf{en$\rightarrow$zh} \\
\midrule
CUNI-AR & 2 & 4 & 2 & -- & -- & -- \\
CUNI-EdUKate & -- & -- & -- & 17 & -- & -- \\
CUNI-MH-v3 & -- & -- & -- & 16 & -- & -- \\
Cohere CAT+ & 16 & 14 & 17 & 19 & 17 & 23 \\
Command A+ & 17 & 16 & 23 & 24 & 18 & 16 \\
DeepSeek V4 Pro & 4 & 5 & 6 & 2 & \textbf{1} & 11 \\
Dubformer & 5 & 12 & 11 & 5 & 3 & 4 \\
GLM 4 - 9B & 18 & 20 & 22 & 23 & 15 & 8 \\
GPT 5.5 & 3 & 9 & 9 & 6 & 4 & 6 \\
GPT OSS 120B & 12 & 15 & 13 & 21 & 12 & 19 \\
GPT OSS 20B & 14 & 23 & 15 & 27 & 21 & 24 \\
Gemini 3.1 Pro & 21 & 26 & 26 & 28 & 23 & 27 \\
Gemma 4 - 31B & 7 & 6 & \textbf{1} & 4 & 9 & 10 \\
Gemma 4 - 4B & 10 & 2 & 19 & 12 & 8 & 5 \\
GoogleTranslate & 11 & \textbf{1} & 20 & 9 & 6 & 3 \\
HW-TSC & -- & 11 & -- & -- & -- & 25 \\
Lumen & -- & 8 & 8 & 11 & 10 & 15 \\
Ministral 3 - 14B & 20 & 24 & 24 & 25 & 14 & 21 \\
Mistral Medium 3.5 & 9 & 7 & 12 & 14 & 5 & 7 \\
Qingqiu-MT-9B & 6 & 21 & 4 & 8 & 19 & 13 \\
Qwen 3.5 - 9b & 15 & 19 & 21 & 26 & 22 & 26 \\
Qwen 3.6 - 27b & 13 & 17 & 16 & 20 & 20 & 17 \\
SCIR-TG-MT & -- & -- & -- & -- & -- & \textbf{1} \\
SRPOL & -- & -- & 18 & 18 & -- & 20 \\
SalamandraTA & 8 & 22 & 14 & 10 & 16 & 9 \\
TRIVE & \textbf{1} & 3 & 5 & 3 & 2 & 12 \\
Tiny Aya Global & -- & 18 & 27 & 22 & 13 & 18 \\
Tower 9B & 19 & 25 & 25 & 13 & 11 & 2 \\
UvA-MT & -- & -- & 10 & 15 & -- & 22 \\
VoxNexus\_V1 & -- & 13 & 7 & \textbf{1} & -- & -- \\
Wayfinder & -- & 10 & 3 & 7 & 7 & 14 \\
\bottomrule
\end{tabular}%
}
\end{table*}

\end{CJK*}
\end{document}